\documentclass[10pt,twocolumn,letterpaper]{article}
\pdfoutput = 1
\usepackage{cvpr}
\usepackage{times}
\usepackage{epsfig}
\usepackage{graphicx}
\usepackage{amsmath}
\usepackage{amssymb}
\usepackage{mathtools}
\usepackage{algorithm}
\usepackage{algpseudocode}
\usepackage{booktabs}
\usepackage{subcaption}

\usepackage{float}

\usepackage[pagebackref=true,breaklinks=true,letterpaper=true,colorlinks,bookmarks=false]{hyperref}

\cvprfinalcopy 

\def\cvprPaperID{2495} 

\ifcvprfinal\fi
\begin{document}
\newcommand{\new}{{\color{red}\textbf{New!}}\,}
\newcommand{\acknowledge}[1]{\textbf{#1}\hspace{0.3em}}
\title{G-CNN: an Iterative Grid Based Object Detector}
\author{Mahyar Najibi \hspace{2em} Mohammad Rastegari \hspace{2em} Larry S. Davis \\
University of Maryland, College Park\\
{\tt\small\{najibi,mrastega\}@cs.umd.edu \hspace{2em} lsd@umiacs.umd.edu}} 

\maketitle
\thispagestyle{empty}

\begin{abstract}
  
  We introduce G-CNN, an object detection technique based on CNNs which works without proposal algorithms. G-CNN starts with a multi-scale grid of fixed bounding boxes. We train a regressor to move and scale elements of the grid towards objects iteratively. G-CNN models the problem of object detection as finding a path from a fixed grid to boxes tightly surrounding the objects. G-CNN with around 180 boxes in a multi-scale grid performs comparably to Fast R-CNN which uses around 2K bounding boxes generated with a proposal technique. This strategy makes detection faster by removing the object proposal stage as well as reducing the number of boxes to be processed.

\end{abstract}


\section{Introduction}

Object detection, \ie the problem of finding the locations of objects and determining their categories, is an intrinsically more challenging problem than classification since it includes the problem of object localization. The recent and popular trend in object detection uses a pre-processing step to find a candidate set of bounding-boxes that are likely to encompass the objects in the image. This step is referred to as the bounding-box proposal stage. The proposal techniques are a major computational bottleneck in state-of-the-art object detectors \cite{girshick2015fast}. There have been attempts \cite{redmon2015you,lenc2015rcnn} to take this pre-processing stage out of the loop but they lead to performance degradations.   

\begin{figure}[t]
\centering
  \includegraphics[width=19pc]{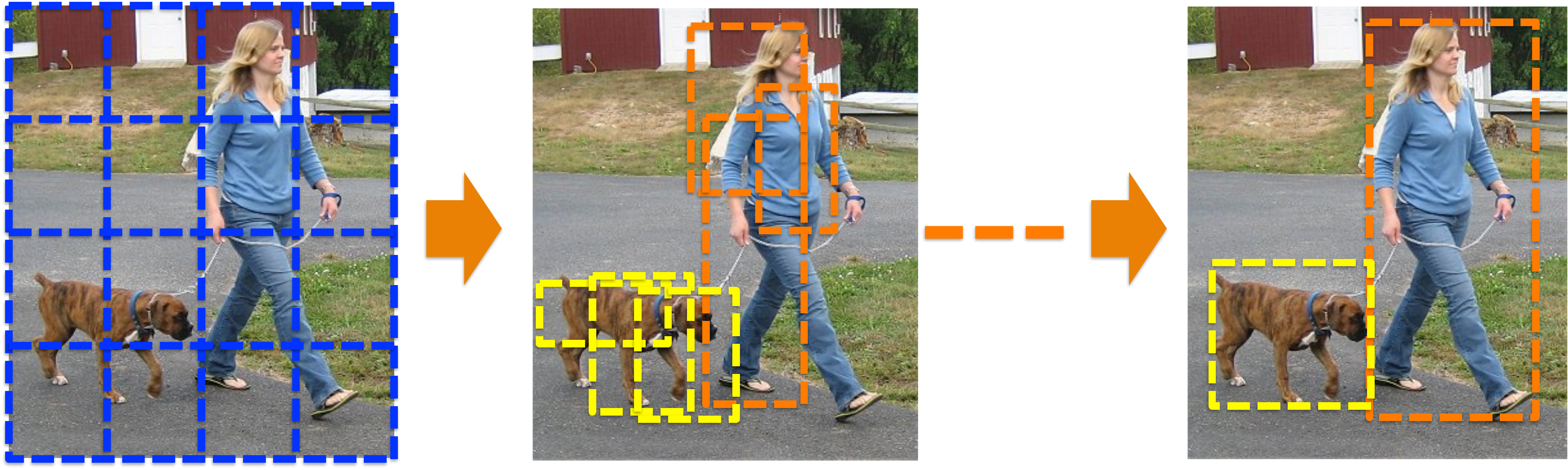}
\caption{\small This figure shows a schematic illustration of our iterative algorithm "G-CNN". It starts with a multi-scale regular grid over the image and iteratively updates the boxes in the grid. Each iteration pushes the boxes toward the objects of interest in the image while classifying their category.}
\label{fig:teaser}
\end{figure}

We show that without object proposals, we can achieve detection rates similar to state-of-the-art performance in object detection. Inspired by the iterative optimization in \cite{carreira2015human}, we introduce an iterative algorithm that starts with a regularly sampled multi-scale grid of boxes in an image and updates the boxes to cover and classify objects. One step regression can-not handle the non-linearity of the mapping from a regular grid to boxes containing objects. Instead, we introduce a piecewise regression model that can learn this non-linear mapping through a few iterations. Each step in our algorithm deals with an easier regression problem than enforcing a direct mapping to actual target locations.   

Figure \ref{fig:teaser} depicts an overview of our algorithm. Initially, a multi-scale regular grid is superimposed on the image. For visualization we show a grid of non-overlapping, but in actuality the boxes do overlap. During training, each box is assigned to a ground-truth object by an assignment function based on intersection over union with respect to the ground truth boxes. Subsequently, at each training step, we regress boxes in the grid to move themselves towards the objects in the image to which they are assigned. At test time, for each box at each iteration, we obtain confidence scores over all categories and update its location with the regressor trained for the currently most probable class.

Our experimental results show that G-CNN achieves the state-of-the-art results obtained by Fast-RCNN on PASCAL VOC datasets without computing bounding-box proposals. Our method is about $5X$ faster than Fast R-CNN for detection.  

\section{Related Work}

\textbf{Prior to CNN:} For many years the problem of object detection was approached by techniques involving sliding window and classification \cite{vedaldi2009multiple,torralba2004contextual}. Lampert \etal \cite{lampert2008beyond} proposed an algorithm that goes beyond sliding windows and was guaranteed to reach the global optimal bounding box for an SVM-based classifier. Implicit Shape Models \cite{leibe2006implicit,maji2009object} eliminated sliding window search by relying on key-parts of an image to vote for a consistent bounding box that covers an object of interest. Deformable Part-based Models \cite{felzenszwalb2008discriminatively} employed an idea similar to Implicit Shape Models, but proposed a direct optimization via latent variable models and used dynamic programming for fast inference. Several extension of DPMs emerged \cite{felzenszwalb2010cascade,azizpour2012object} until the remarkable improvements due to the convolutional neural networks was shown by \cite{girshick2014rich}.

\textbf{CNN age:} Deep convolutional neural networks (CNNs) are the state-of-the-art image classifiers and successful methods have been proposed based on these networks \cite{krizhevsky2012imagenet}. Driven by their success in image classification, Girshick \etal proposed a multi-stage object detection system, known as R-CNN \cite{girshick2014rich}, which has attracted great attention due to its success on standard object detection datasets. 

To address the localization problem, R-CNN relies on advances in object proposal techniques. Recently, proposal algorithms have been developed which avoid exhaustive search of image locations \cite{uijlings2013selective,zitnick2014edge}. R-CNN uses these techniques to find bounding boxes which include an object with high probability. Next, a standard CNN is applied as feature extractor to each proposed bounding box and finally a classifier decides which object class is inside the box.

The main drawback of R-CNN is the redundancy in computing the features. Generally, around 2K proposals are generated; for each of them, the CNN is applied independently to extract features. To alleviate this problem, in SPP-Net \cite{he2014spatial} the convolutional layers of the network are applied only once for each image. Then, the features of each region of interest are constructed by pooling the global features which lie in the spatial support of the region. However, learning is limited to fine-tuning the weights of fully connected layers. This drawback is addressed in Fast-RCNN \cite{girshick2015fast} in which all parameters are learned by back propagating the errors through the augmented pooling layer and packing all stages of the system, except generation of the object proposals, into one network. 


The generation of object proposals, in CNN-based detection systems has been regarded as crucial. However, after proposing Fast-RCNN, this stage became the bottleneck. To make the number of object proposals smaller, Multibox\cite{erhan2014scalable} introduced a proposal algorithm that outputs 800 bounding boxes using a CNN. This increases the size of the final layer of the CNN to 4096x800x5 and introduces a large set of additional parameters. Recently, Faster-RCNN \cite{ren2015faster} was proposed, which decreased the number of parameters; however it needs to start from thousands of anchor points to propose 300 boxes.

In addition to classification, using a regressor for object detection has been also studied previously. Before proposing R-CNN, Szegedy \etal \cite{szegedy2013deep}, modeled object detection as a regression problem and proposed a CNN-based regression system. More recently, AttentionNet \cite{yoo2015attentionnet} is a single category detection that detects a single object inside an image using iterative regression. For multiple objects, the model is applied as a proposal algorithm to generate thousands of proposals and then is re-applied iteratively on each proposal for single category detection, which makes detection inefficient. 

Although R-CNN and its variants attack the problem using a classification approach, they employ regression as a post-processing stage to refine the localization of the proposed bounding boxes.

The importance of the regression stage has not received as much attention as improving the object proposal stage for more accurate localization. The necessity of an object proposal algorithm in CNN based object detection systems has recently been challenged by Lenc \etal \cite{lenc2015rcnn}. Here, the proposals are replaced by a fixed set of bounding boxes. A set with a distribution derived from an object proposal method is selected using a clustering technique. However, for achieving comparable results, even more boxes need to be used compared to R-CNN. Another recent attempt for removing the proposal stage is Redmon \etal \cite{redmon2015you} which conducts object detection in a single shot. However, the considerable gap between the best detection accuracy of these systems and systems with an explicit proposal stage suggests that the identification of good object proposals is critical to the success of these CNN based detection systems.

\section{G-CNN Object Detector}

\subsection{Network structure}
\begin{figure*}
\centering
  \includegraphics[width=0.65\textwidth]{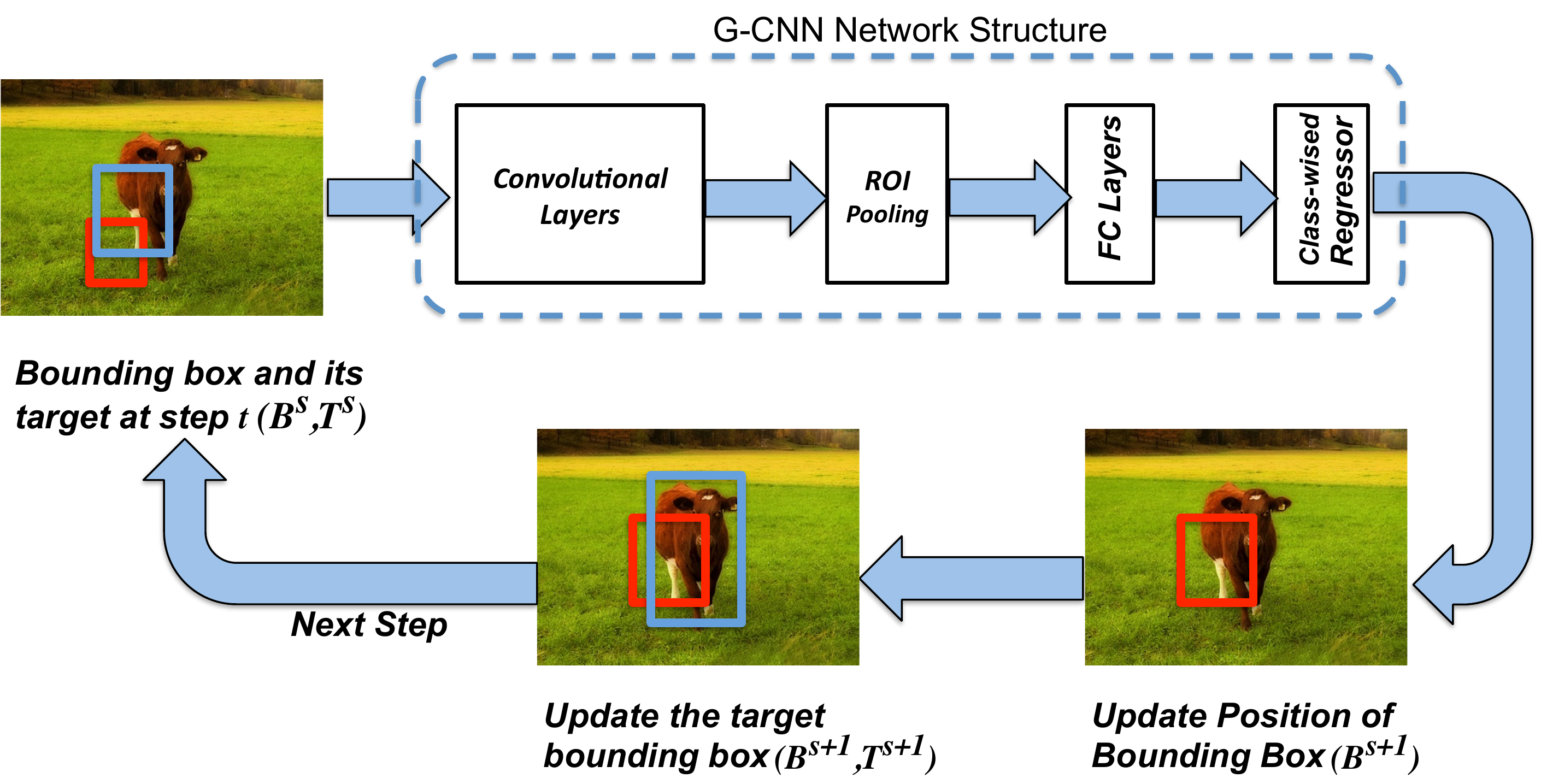}
\caption{\small Structure of G-CNN regression network as well as an illustration of the idea behind the iterative training approach. The bounding box at each step is shown by the blue rectangle and its target is represented by a red rectangle. The network is trained to learn the path from the initial bounding box to its assigned target iteratively.}
\label{fig:structure}
\end{figure*}

G-CNN trains a CNN to move and scale a fixed multi-scale grid of bounding boxes towards objects. The network architecture for this regressor is shown in Figure \ref{fig:structure}. The backbone of this architecture can be any CNN network (\eg AlexNet \cite{krizhevsky2012imagenet}, VGG \cite{simonyan2014very}, \etc). As in Fast R-CNN and SPP-Net, a spatial region of interest (ROI) pooling layer is included in the architecture after the convolutional layers. Given the location information of each box, this layer computes the feature for the box by pooling the global features that lie inside the ROI. After the fully connected layers, the network ends with a linear regressor which outputs the change in the location and scale of each current bounding box, conditioned on the assumption that the box is moving towards an object of a class.

\subsection {Training the network}

Despite the similarities between the Fast R-CNN and G-CNN architectures, the training goals and approaches are different. G-CNN defines the problem of object detection as an iterative search in the space of all possible bounding boxes. G-CNN starts from a fixed multi-scale spatial pyramid of boxes. The goal of learning is to train the network so that it can move this set of initial boxes towards the objects inside the image in $S$ steps iteratively. This iterative behaviour is essential for the success of the algorithm. The reason is the highly non-linear search space of the problem. In other words, although learning how to linearly regress boxes to far away targets is unrealistic, learning small changes in the search space is tractable. Section \ref{sec:step_wised} shows the importance of this step-wise training approach.

\subsubsection{Loss function}

G-CNN is an iterative method that moves bounding boxes towards object locations in $S_{train}$ steps. For this reason, the loss function is defined not only over the training samples but also over the iterative steps.

More formally, let $\mathcal{B}$ represent the four-dimensional space of all possible bounding boxes represented by the coordinates of their center, their width, and height. $\mathbf{B}_i \in \mathcal{B}$  is the $i$'th training bounding box. We use the superscript $ 1 \leq s \leq S_{train}$ to denote the variables in step 's' of the G-CNN training, \ie $\mathbf{B}_i^s$ is the position of the $i$'th training bounding box in step $s$.

During training, each bounding box with an IoU higher than a small threshold (0.2) is assigned to one of the ground truth bounding boxes inside its image. The following many-to-one function, $\mathcal{A}$, is used for this assignment.

\begin{equation}
 \label{eq:assignment}
 \mathcal{A}(\mathbf{B}_i^s) =
 arg\max_{\mathbf{G} \in \mathcal{G}_i} IoU(\mathbf{B}_i^1, \mathbf{G})
 \end{equation}
where $\mathcal{G}_i = \{ \mathbf{G}_{i1}\in \mathcal{B},\hdots, \mathbf{G}_{in}\in \mathcal{B}\}$, is the set of ground truth bounding boxes which lie in the same image as $\mathbf{B}_i$. IoU is the intersection over union measure. Note that $\mathbf{B}_i^1$ represents the position of the $i$'th bounding box in the initial grid. In other words, for each training bounding box, the assignment is done in the initial training step and is not changed during the training.

Since regressing the initial training bounding boxes to their assigned ground truth bounding box can be highly non-linear, we tackle the problem with a piece-wise regression approach. At step $s$, we solve the problem of regressing $\mathbf{B}_i^s$ to a target bounding box on the path from $\mathbf{B}_i^s$ to its assigned ground truth box. The target bounding box is moved step by step towards the assigned bounding box until it coincides with the assigned ground truth in step $S_{train}$. The following function is used for defining the target bounding boxes at each step:

\begin{equation}
\label{eq:target}
\Phi(\mathbf{B}_i^s,\mathbf{G}_i^*,s) = \mathbf{B}_i^s + \frac{\mathbf{G}_i^* - \mathbf{B}_i^s}{S_{train}-s + 1}
\end{equation}

where $\mathbf{G}_i^*  = \mathcal{A}(\mathbf{B}_i^s)$ represents the assigned ground truth bounding box to $\mathbf{B}_i^s$. That is, at each step, the path from the current representation of the bounding box to the assigned ground truth is divided by the number of remaining steps and the target is defined to be one unit away from the current location.

G-CNN regression network outputs four values for each class, representing the parameterized change for regressing the bounding boxes assigned to that class. Following \cite{girshick2014rich}, a log-scale shift in width and height and a scale invariant translation is used to parametrize the relative change for mapping a bounding box to its assigned target. This parametrization is denoted by $\Delta(\mathbf{B}_i^s,\mathbf{T}_i^s)$, where $\mathbf{T}_i^s$ is the assigned target to $\mathbf{B}_i^s$ computed by \ref{eq:target}.

So the loss function for G-CNN is defined as follows:

\begin{align}
\label{eq:loss}
L(\{\mathbf{B}_i\}_{i=1}^N) =& \sum_{s=1}^{S_{train}} \sum_{i=1}^{N} \big[ I(\mathbf{B}_i^1 \not \in \mathcal{B}_{BG}) \times\\
&L_{reg}(\mathbf{\delta}_{i,l_i}^s - \Delta(\mathbf{B}_i^s,\Phi(\mathbf{B}_i^s,\mathcal{A}(\mathbf{B}_i^s),s)))\big] \nonumber
\end{align}

where $\delta_{i,l_i}^s$ is the four-dimensional parameterized output for class $l_i$ representing the relative change in the representation of bounding box $\mathbf{B}_i^s$. $l_i$ is the class label of the assigned ground truth bounding box to $\mathbf{B}_i$. $L_{reg}$ is the regression loss function. The smooth $l_1$ loss is used as defined in \cite{girshick2015fast}. $I(.)$ is the indicator function which outputs one when its condition is satisfied and zero otherwise. $\mathcal{B}_{BG}$ represents the set of all background bounding boxes.

During training, the representation of bounding box $\mathbf{B}_i$ at step $s$, $\mathbf{B}_i^s$, can be determined based on the actual output of the network by the following update formula:

\begin{equation}\label{eq:update}
\mathbf{B}_i^s = \mathbf{B}_i^{s-1} + \Delta^{-1}(\delta_{i,l_i}^{s-1})
\end{equation}
where $\Delta^{-1}$ projects back the relative change in the position and scale from the defined parametrized space into $\mathcal{B}$. However for calculating \ref{eq:update}, the forward path of the network needs to be evaluated during training, making training inefficient. Instead, we use an approximate update by assuming that in step $s$, the network could learn the regressor for step $s-1$ perfectly. As a result the update formula becomes $\mathbf{B}_i^s = \Phi(\mathbf{B}_i^{s-1},\mathbf{G}_i^*,s-1)$. This update is depicted in Figure \ref{fig:structure}.

\subsubsection{Optimization}
G-CNN optimizes the objective function in \ref{eq:loss} with stochastic gradient descent. Since G-CNN tries to map the bounding boxes to their assigned ground-truth boxes in $S_{train}$ steps, we use a step-wised learning algorithm that optimizes Eq. \ref{eq:loss} step by step. 

To this end, we treat each of the bounding boxes in the initial grid together with its target in each of the steps as an independent training sample \ie for each of the bounding boxes we have $S_{train}$ different training pairs. The algorithm first tries to optimize the loss function for the first step using $N_{iter}$ iterations. Then the training pairs of the second step are added to the training set and training continues step by step. By keeping the samples of the previous steps in the training set, we make sure that the network does not forget what was learned in the previous steps. 

 The samples for the earlier steps are part of the training set for a longer period of time. This choice is made since the earlier steps determine the global search direction and have a greater impact on the chance that the network will find the objects. On the other hand, the later steps only refine the bounding boxes to decrease localization error. Given that the search direction was correct and a good part of the object is now visible in the bounding box, the later steps solve a relatively easier problem.

Algorithm \ref{alg:train} is the method for generating training samples from each bounding box during each G-CNN step.

\begin{algorithm}
\caption{G-CNN Training Algorithm}
\begin{algorithmic}[1]
\Procedure{TrainGCNN}{}
\For {$1\leq c \leq S_{train}$}
\State{TrainTuples $\gets \{ \}$}
    \For{$1\leq s \leq c$}
        \If{$s = 1$}
            \State{$\mathbf{B}^1 \gets$ Spatial pyramid grid of boxes}
            \State{$\mathbf{G}^* \gets \mathcal{A}(\mathbf{B}^1)$ }
        \Else
            \State{$\mathbf{B}^s \gets$ $\mathbf{T}^{s-1}$}
        \EndIf
        \State $\mathbf{T}^s \gets \Phi(\mathbf{B}^s,\mathbf{G}^*,s)$
        \State $\Delta^s \gets \Delta(\mathbf{B}^s,\mathbf{T}^s)$
        \State Add $(\mathbf{B}^s, \Delta^s)$ to TrainTuples
    \EndFor
    \State Train G-CNN $N_{iter}$ iterations with TrainTuples    
\EndFor
\EndProcedure
\end{algorithmic}
\label{alg:train}
\end{algorithm}

\subsection{Fine-tuning}
All models are fine-tuned from pre-trained models on ImageNet. Following \cite{girshick2015fast}, we fine-tune all layers except early convolutional layers (\ie \textit{conv2} and up for AlexNet and \textit{conv3\_1} and up for VGG16). During training, mini-batches of two images are used. At each step of G-CNN, 64 training samples are selected randomly from all possible samples of the image at the corresponding step. 

\subsection{G-CNN Test Network}
\label{sec:gcnn_test}
The G-CNN regression network is trained to detect objects in an iterative fashion from a set of fixed bounding boxes in a multi-scale spatial grid. Likewise at test time, the set of bounding boxes is initialized to boxes inside a spatial pyramid grid. The regressor moves boxes towards objects using the classifier score to determine which class regressor to apply to update the box. The detection algorithm is presented in Algorithm \ref{al:detect}.

\begin{figure}
\centering
  \includegraphics[width=0.29\textwidth]{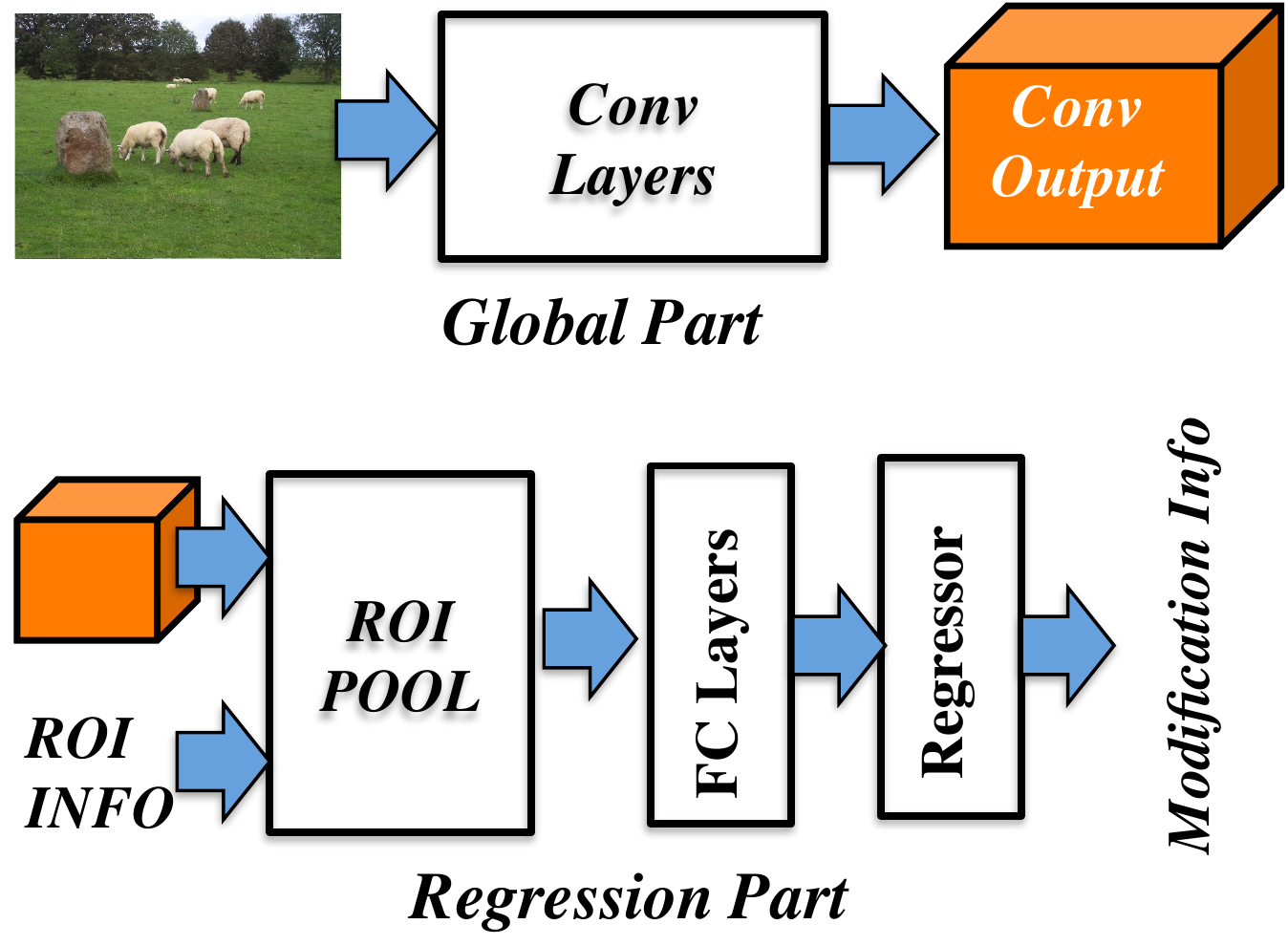}
\caption{\small Decomposition of the G-CNN network into global (upper) and regression part (lower) for detection after the training phase. Global part is run only once to extract global features but regression part is run at every iteration. This leads to a considerable speed up at test time.}
\label{fig:test_nets}
\end{figure}

During the detection phase, G-CNN is run $S_{test}$ times. However, like SPP-Net and Fast R-CNN there is no need to compute activations for all layers at every iteration. During test time, we decompose the network into global and regression parts as depicted in Figure.  \ref{fig:test_nets}. The global net contains all convolutional layers of the network. On the other hand, the regression part consists of the fully connected layers and the regression weights. The input to the global net is the image and the forward path is computed only once for each image, outside the detection loop of Algorithm \ref{al:detect}. Inside the detection loop, we only operate the regression network, which takes the outputs of the last layer of the global net as input and produces the bounding box modifications. 

\begin{algorithm}[b]
\caption{Detection algorithm}
\begin{algorithmic}[1]
\State Let f(.) be the feed-forward G-CNN regression network
\State Let c(.) be the classifier function
\Procedure{Detect}{}
\State $\mathbf{B}^1 \gets $ Spatial pyramid grid of boxes
\For{$1\leq s \leq S_{test}$}
    \State $l \gets c(\mathbf{B}^{s})$
    \State $\delta_l^s \gets f(\mathbf{B}^{s})$
    \State $\mathbf{B}^{s+1} \gets \mathbf{B}^{s} + \Delta^{-1}(\delta_l^s)$ 
\EndFor
\State Output $\mathbf{B}^{S_{test}+1}$
\EndProcedure
\end{algorithmic}
\label{al:detect}
\end{algorithm}

This makes the computational cost of the algorithm comparable to Fast R-CNN (without considering the object proposal stage of Fast R-CNN). The global net is called once in both Fast R-CNN and G-CNN. Afterward, Fast R-CNN does $N_{proposal}$ forward calculations of the regression network (where $N_{proposal}$ is the number of generated object proposals for each image). G-CNN, on the other hand, does this forward calculation $S_{test}\times N_{grid}$ times (where $N_{grid}$ is the number of bounding boxes in the initial grid). In section \ref{sec:voc}, we show that for $S_{test} = 5$ and $N_{grid}  \sim 180$, G-CNN achieves comparable results to Fast R-CNN which uses $N_{proposal} \sim 2K$ object proposals.

\section{Experiments}
\subsection{Experimental Setup}
\label{sec:exp_setup}
We report results on the Pascal VOC 2007 and Pascal VOC 2012 datasets. The performance of G-CNN is evaluated with AlexNet \cite{krizhevsky2012imagenet} as a small and VGG16 \cite{simonyan2014very} as a very deep CNN structure. Following \cite{girshick2014rich}, we scale the shortest side of the image to 600 pixels not allowing the longer side of the image to be more than 1000 pixels. However, we always maintain the aspect ratio of the image, so the shortest side might include fewer than 600 pixels. Each model is pre-trained with weights learned from the imagenet dataset.

In all the experiments, the G-CNN regression network is trained on an initial overlapping spatial pyramid with [2,5,10] scales (\ie the bounding boxes in the coarsest level are $(im_{width}/2,im_{height}/2)$ pixels \etc). During training, we used [0.9,0.8,0.7] overlap for each spatial scale respectively. By overlap of $\alpha$ we mean that the horizontal and vertical strides are $width_{cell}*(1-\alpha)$ and $height_{cell}*(1-\alpha)$ respectively. However, during test time, as will be shown in the following sections, overlaps of [0.7,0.5,0] (non-overlapping grid at the finest scale) is sufficient to obtain results comparable to Fast R-CNN. This leads to a grid of almost 180 initial boxes at test time. The G-CNN regression network is trained for $S=3$ iterative steps. According to our experiments, no substantial improvement is achieved by training the network for a larger number of steps.   

\subsection{Results on VOC datasets}
\label{sec:voc}
The goal of G-CNN is to replace object proposals with a fixed multi-scale grid of boxes. To evaluate this, we fix the classifier in Algorithm \ref{al:detect} to the Fast R-CNN classifier and compare our results to the original Fast R-CNN with selective search proposal algorithm.

\subsubsection{VOC 2007 dataset}
Table \ref{tab:voc07_alexnet} compares the mAP between G-CNN and Fast R-CNN on the VOC2007 \textit{test} set. AlexNet is used as the basic CNN for all methods and models are trained on VOC2007 \textit{trainval} set. G-CNN(3) is our method with three iterative steps during test time. In this version, we used the same grid overlaps used during training. This leads to a set of around 1500 initial boxes. G-CNN(5) is our method when we increase the number of steps at test time to 5 but reduce the overlaps to [0.7,0.5,0] (see \ref{sec:exp_setup}). This leads to around 180 boxes per image. According to the result, 180 boxes is enough for G-CNN to surpass the performance of Fast R-CNN, which uses around 2K selective search proposed boxes. In the remainder of this paper, we use G-CNN to refer to the G-CNN(5) version of our method.

Table \ref{tab:voc07_vgg} shows mAP for various methods trained on VOC2007 trainval set and tested on VOC2007 test set. All methods used VGG16. The results validate our claim that G-CNN effectively moves its relatively small set of boxes toward objects. In other words, there seems to be no advantage to employing the larger set of selective search proposed boxes for detection in this dataset.

\begin{table*}[htbp]

\caption{\small Average Precision on VOC 2007 test data. Both Fast R-CNN and our methods use AlexNet CNN structure. Models are trained using VOC 2007 trainval set.}
    \centering
    \resizebox{\textwidth}{!}{\begin{tabular}{l|c c c c c c c c c c c c c c c c c c c c| c}
    \toprule
    \bf{VOC 2007} & aero & bike & bird & boat & bottle & bus & car & cat & chair & cow & table & dog & horse & mbike & person & plant & sheep & sofa & train & tv & mAP\\
    \midrule
    FR-CNN \cite{girshick2015fast}& \bf{66.4}&\bf{71.6}&\bf{53.8}&43.3&24.7&69.2&\bf{69.7}&\bf{71.5}&31.1&\bf{63.4}&59.8&62.2&73.1&65.9&\bf{57}&26&\bf{52}&56.4&67.8&\bf{57.7}&57.1\\
    \hline
    G-CNN(3) [ours]& 63.2&68.9&51.7&41.8&\bf{27.2}&69.1&67.7&69.2&31.8&60.6&\bf{60.8}&63.9&\bf{75.5}&67.3&54.9&26.1&51.2&\bf{57.2}&69.6&56.8&56.7\\
    
    G-CNN(5) [ours]&
    65&68.5&52&\bf{44.9}&24.5&\bf{69.3}&69.6&68.9&\bf{34.6}&60.3&58.1&\bf{64.6}&75.1&\bf{70.5}&55.2&\bf{28.5}&50.7&56.8&\bf{70.2}&56.1&\bf{57.2}\\
    \hline
    \end{tabular}}
    \label{tab:voc07_alexnet}
\end{table*}

\begin{table*}[htbp]

\caption{\small Average Precision on VOC 2007 Test data. All reported methods used VGG16. Models are trained using VOC 2007 trainval set.}
    \centering
    \resizebox{\textwidth}{!}{\begin{tabular}{l|c c c c c c c c c c c c c c c c c c c c| c}
    \toprule
    \bf{VOC 2007} & aero & bike & bird & boat & bottle & bus & car & cat & chair & cow & table & dog & horse & mbike & person & plant & sheep & sofa & train & tv & mAP\\
    \midrule
    SPPnet BB\cite{he2014spatial} & 
    73.9&
    72.3 &  62.5 &  51.5&    44.4&    74.4 &  73.0&   74.4 &   42.3 &   73.6 &  57.7 &  70.3    &74.6&     74.3&     54.2&    34.0 &    56.4    &56.4 &  67.9&   \bf{73.5}& 63.1\\

    R-CNN BB\cite{girshick2016region} & 73.4 & 77.0 &  63.4  & 45.4 & \bf{44.6} & 75.1 &78.1   & 79.8    &40.5   & \bf{73.7}  & 62.2  & 79.4 &   78.1 & 73.1     & 64.2 &\bf{35.6} &\bf{66.8}  &  67.2 &  70.4 & 71.1 & 66.0 \\
    
    FR-CNN\cite{girshick2015fast}& \bf{74.5} &
\bf{78.3}&   \bf{69.2}&   \bf{53.2}&    36.6 &   77.3 &  78.2&   \bf{82.0} &   40.7&    72.7&   \bf{67.9}&   \bf{79.6}    &79.2&     73.0 &    \bf{69.0}&    30.1&     65.4&    \bf{70.2}&   \bf{75.8}&   65.8&
\bf{66.9}\\
\hline
    G-CNN[ours]&
    68.3&77.3&68.5&52.4&38.6&\bf{78.5}&\bf{79.5}&81&\bf{47.1}&73.6&64.5&77.2&\bf{80.5}&\bf{75.8}&66.6&34.3&65.2&64.4&75.6&66.4&66.8\\
    \hline
    
    \end{tabular}}
    \label{tab:voc07_vgg}
\end{table*}

\subsubsection{VOC 2012 dataset}
The mAP for VOC2012 dataset is reported in Table \ref{tab:voc2012_vgg}. All methods use VGG16 as their backbone. Methods are trained on trainval set and tested on the VOC2012 test set. The results of our method are obtained using the "comp4" evaluation server with the parameters mentioned in \ref{sec:exp_setup} and the results of other methods are obtained from their papers.



    
    

\begin{table*}[!ht]
\caption{\small Average Precision on VOC2012 test data. All reported methods used VGG16. The training set for each image is mentioned in the second column (12 stands for VOC2012 trainval, \textbf{07+12} represents the union of the trainval of VOC2007 and VOC2012, and \textbf{07++12} is the union of VOC 2007 trainval, VOC 2007 test and VOC 2012 trainval. The * emphasises that our method is trained on fewer data compared to FR-CNN trained on 07++12 training data)}
    \centering
    \resizebox{\textwidth}{!}{\begin{tabular}{l|c|c c c c c c c c c c c c c c c c c c c c| c}
    \toprule
    \bf{VOC 2012}& train & aero & bike & bird & boat & bottle & bus & car & cat & chair & cow & table & dog & horse & mbike & person & plant & sheep & sofa & train & tv & mAP\\
    \midrule

    R-CNN BB\cite{girshick2016region} & 12 & 79.6
&72.7 &  61.9  & 41.2 &   41.9  &  65.9 &  66.4 &  84.6&    38.5&    67.2&   46.7 &  82.0&    74.8 &    76.0    & 65.2&    35.6&     65.4&    54.2&   67.4&   60.3&
62.4 \\
    
    YOLO\cite{redmon2015you}&12&71.5&64.2&54.1&35.3&23.3&61.0&54.4&78.1&35.3&56.9&40.9&72.4&68.6&68.0&62.5&26.0&51.9&48.8&68.7&47.2& 54.5\\
    FR-CNN\cite{girshick2015fast}&12 &80.3&
74.7 &  66.9 &  46.9    &37.7&    73.9&   68.6&   87.7&    41.7 &   71.1&   51.1 &  86.0    &77.8&     79.8&     69.8&    32.1&     65.5&    63.8&   76.4&   61.7& 65.7\\
    FR-CNN\cite{girshick2015fast}&07++12 &82.3&
78.4 &  70.8 &  52.3&
38.7&
77.8 &  71.6 &  89.3&    44.2 &   73.0 &  55.0 &  87.5  &  80.5     &80.8&     72.0&
35.1&
68.3&    65.7&   80.4&   64.2&
68.4\\

\hline
G-CNN [ours] &12 &82&74&
68.2&
49.5&
38.9&
74.4&
68.9&
85.4&
40.6&
70.9&
50&
85.5&
77&
77.4&
67.9&
33.7&
67.6&
60&
77.6&
60.8&
65.5\\

G-CNN [ours] &07+12 & 82&
76.1&
69.3&
49.9&
40.1&
75.2&
69.5&
86.3&
42.3&
72.3&
50.8&
84.7&
77.8&
77.2&
68&
38.1&
68.4&
59.8&
79.1&
61.9&
66.4*\\
    \hline
    
    \end{tabular}}
    \label{tab:voc2012_vgg}
\end{table*}


G-CNN obtains almost the same result as Fast R-CNN when both methods are trained on VOC 2012 trainval. Although in this table the best-reported mAP for Fast RCNN is slightly higher than G-CNN, it should be noted that unlike G-CNN, Fast R-CNN used the VOC 2007 test set as part of its training. It is worth noting that all methods except YOLO use proposal algorithms with high computational complexity. Compared to YOLO, which does not use object proposals, our method has a considerably higher mAP. To the best of our knowledge, this is the best-reported result among methods without an object proposal stage.

\subsection{Stepwise training matters}
\label{sec:step_wised}
G-CNN uses a stepwise training algorithm and defines its loss function with this goal. In this section, we investigate the question of how important this stepwise training is and whether it can be replaced by a simpler, single step training approach. 

To this end, we compare G-CNN with two simpler iterative approaches in table \ref{tab:diff_methods}. First we consider the iterative version of Fast R-CNN (\emph{IF-RCNN}). In this method, we use the regressor trained with Fast R-CNN in our iterative framework. Clearly, this regressor was not designed for grid-based object detection, but for small post-refinement of proposed objects.

Also, we consider a simpler algorithm for training the regressor for a grid-based detection system. Specifically, we collect all training tuples created in different steps of G-CNN and train our regressor in one step on this training set. So the only difference between G-CNN and this method is stepwise training. We call this method \emph{1Step-Grid}.

\begin{figure}
\centering
\includegraphics[width=0.27\textwidth]{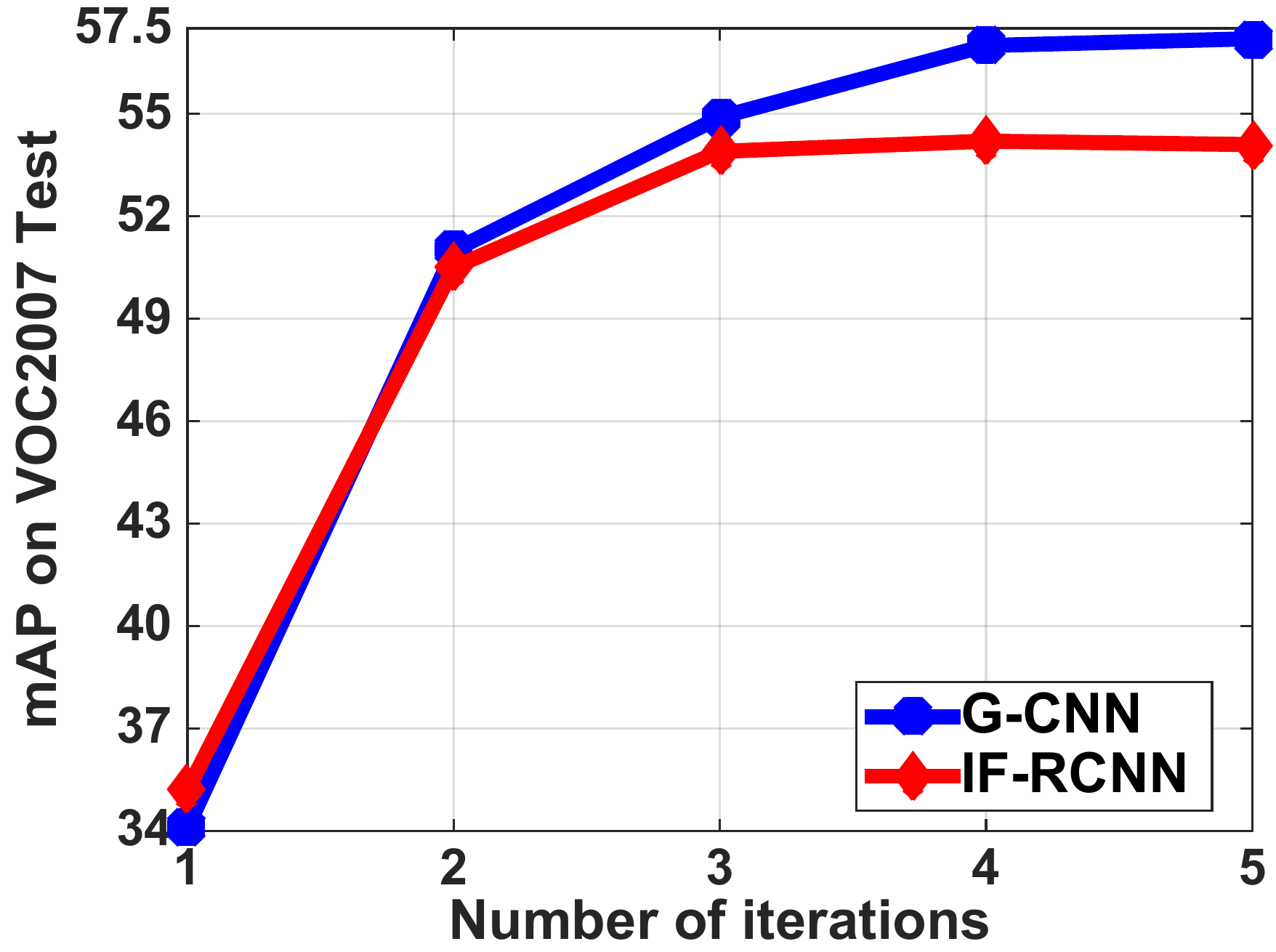}
\caption{\small Mean average precision on VOC2007 test set vs. number of regression steps for G-CNN and IF-RCNN. Both methods use AlexNet and trained on VOC2007 trainval.}
\label{fig:iterative_comparision}
\end{figure}

All methods are trained on VOC 2007 trainval set and tested on VOC 2007 test set and AlexNet is used as the core CNN structure. All methods are applied five iterations during test time to the same initial grid. Table \ref{tab:diff_methods} shows the comparison among the methods and Figure \ref{fig:iterative_comparision} compares IF-RCNN and G-CNN for different numbers of iterations.

\begin{table*}[!ht]
\caption{\small Comparison among different strategies for grid-based object detection trained on VOC2007 trainval. All methods used AlexNet.}
    \centering
    \resizebox{\textwidth}{!}{\begin{tabular}{l|c c c c c c c c c c c c c c c c c c c c| c}
    \toprule
    \bf{VOC 2007} & aero & bike & bird & boat & bottle & bus & car & cat & chair & cow & table & dog & horse & mbike & person & plant & sheep & sofa & train & tv & mAP\\
    \midrule
    IF-RCNN & 
    51.3&67.1&51.6&33.7&\bf{26.2}&67.8&66.3&\bf{70.3}&31.5&56.3&55.9&62.6&74.7&64.6&\bf{55.6}&22.2&46.5&54.3&67.4&55&54.1\\

    1Step-Grid&59.6&63.3&\bf{52.4}&40.2&20.9&68.1&67.1&68.6&29.7&59.6&\bf{62.1}&63&70.7&64&53.2&23.4&50.1&56&63.5&53.9&54.5\\
\hline
    G-CNN [ours] & 
    \bf{65}&\bf{68.5}&52&\bf{44.9}&24.5&\bf{69.3}&69.6&68.9&\bf{34.6}&\bf{60.3}&58.1&\bf{64.6}&\bf{75.1}&\bf{70.5}&55.2&\bf{28.5}&\bf{50.7}&\bf{56.8}&\bf{70.2}&\bf{56.1}&\bf{57.2}\\
    
    \hline
    
    \end{tabular}}
    \label{tab:diff_methods}
\end{table*}

The results show that step-wise training is crucial to the success of G-CNN. Even though the training samples are the same for G-CNN and 1Step-Grid, G-CNN outperforms it by a considerable margin.

\subsection{Analysis of the detection results}
G-CNN removes the proposal stage from CNN-based object detection networks. Since the object proposal stage is known to be important for achieving good localization in CNN-based techniques, we compare the localization of G-CNN with Fast R-CNN.

\begin{figure}
    \centering
    \begin{subfigure}[b]{0.155\textwidth}
        \includegraphics[width=\textwidth]{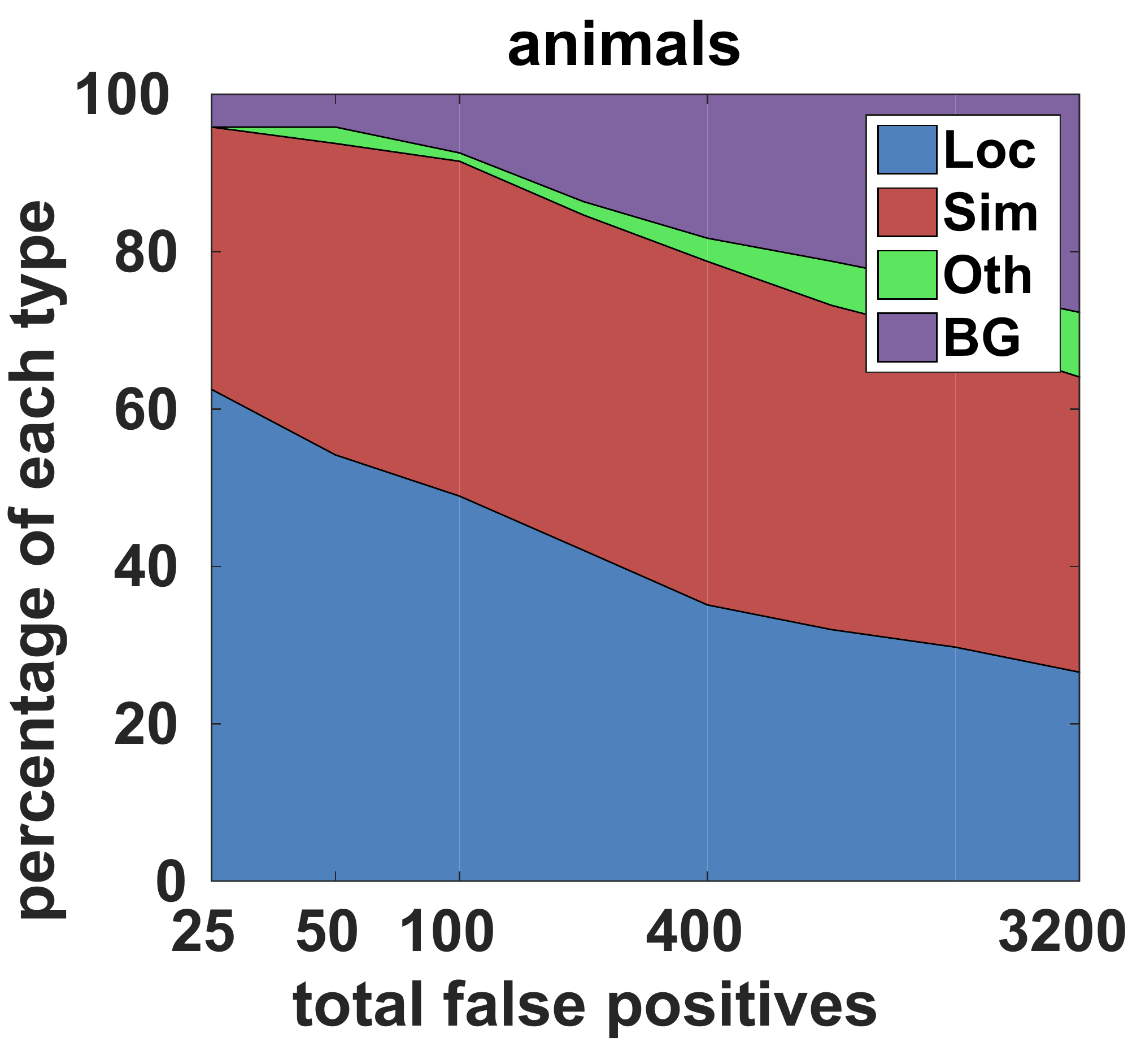}
    \end{subfigure}
    \begin{subfigure}[b]{0.155\textwidth}
        \includegraphics[width=\textwidth]{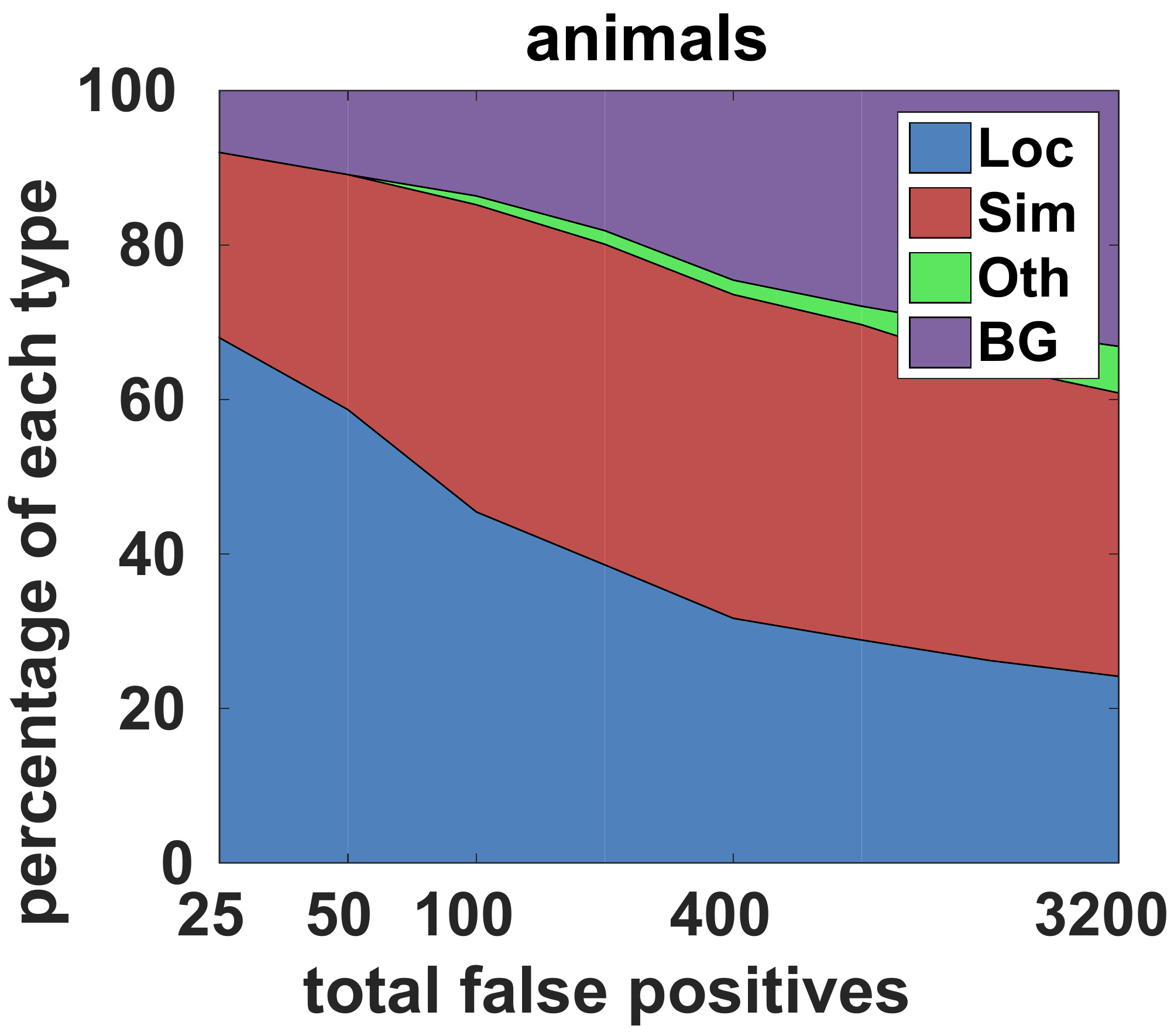}
    \end{subfigure}
    \begin{subfigure}[b]{0.155\textwidth}
        \includegraphics[width=\textwidth]{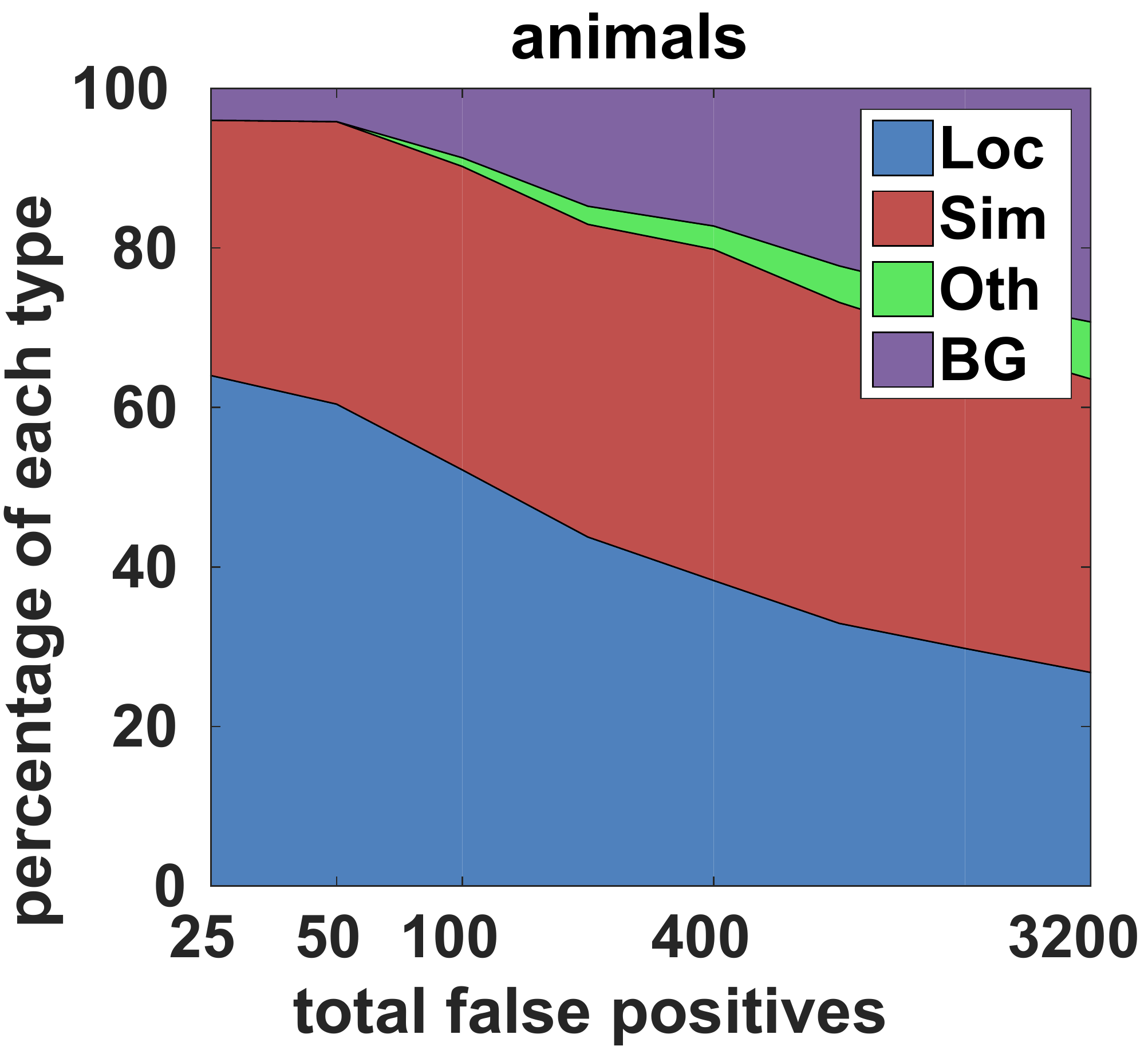}
    \end{subfigure}
    
    \begin{subfigure}[b]{0.155\textwidth}
        \includegraphics[width=\textwidth]{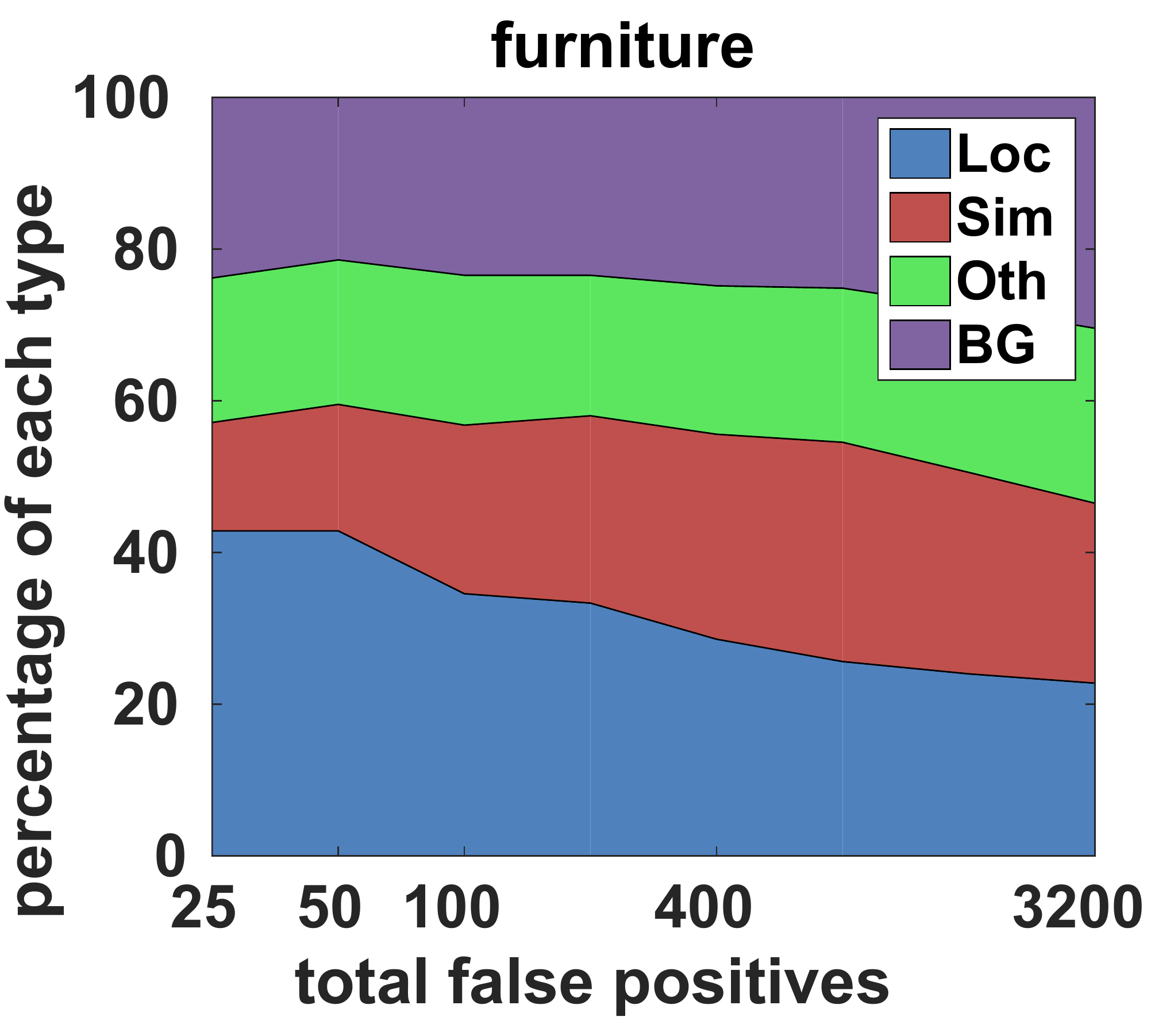}
        \caption{G-CNN}
    \end{subfigure}
    \begin{subfigure}[b]{0.155\textwidth}
        \includegraphics[width=\textwidth]{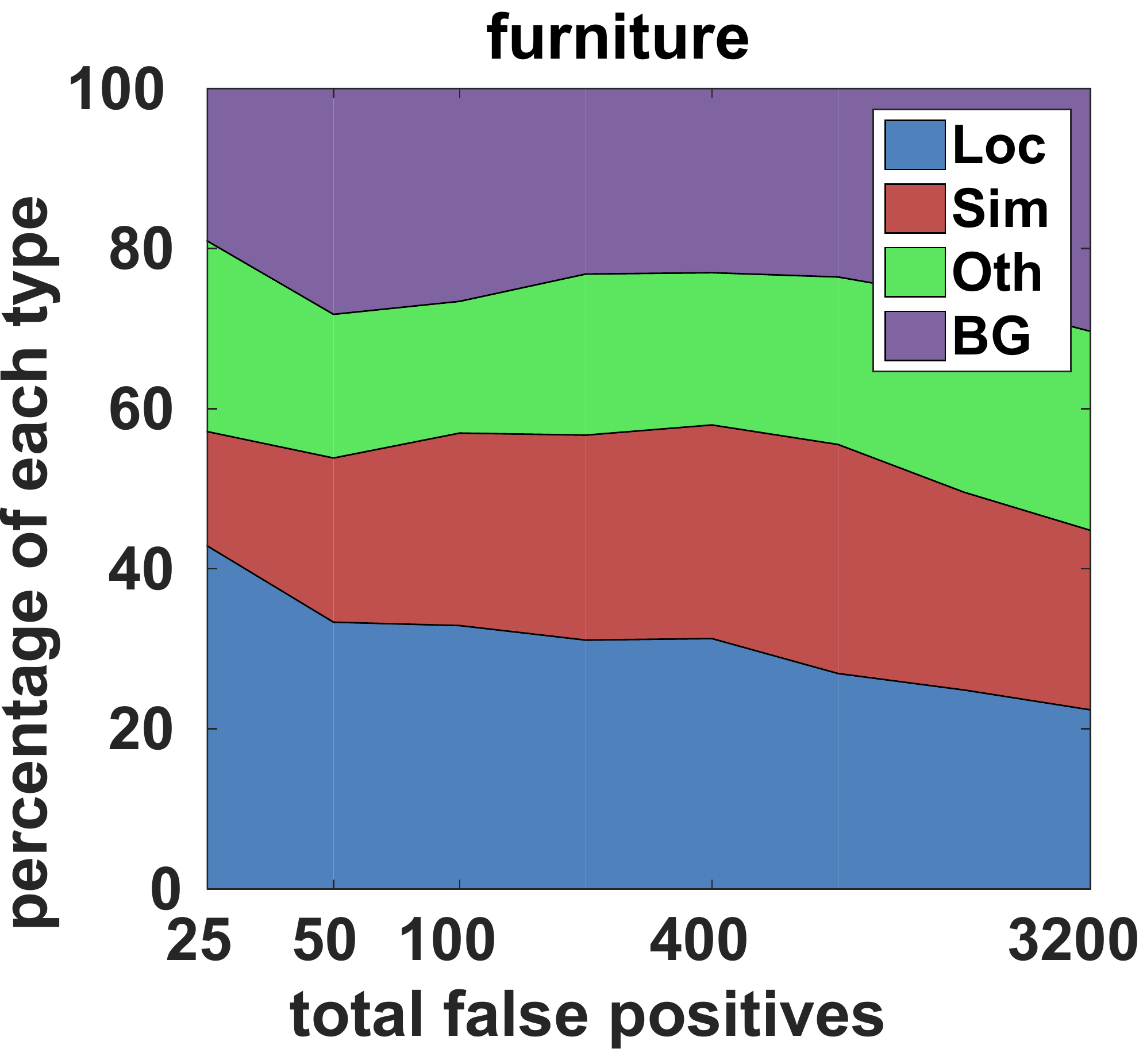}
        \caption{Fast R-CNN}
    \end{subfigure}
    \begin{subfigure}[b]{0.155\textwidth}
        \includegraphics[width=\textwidth]{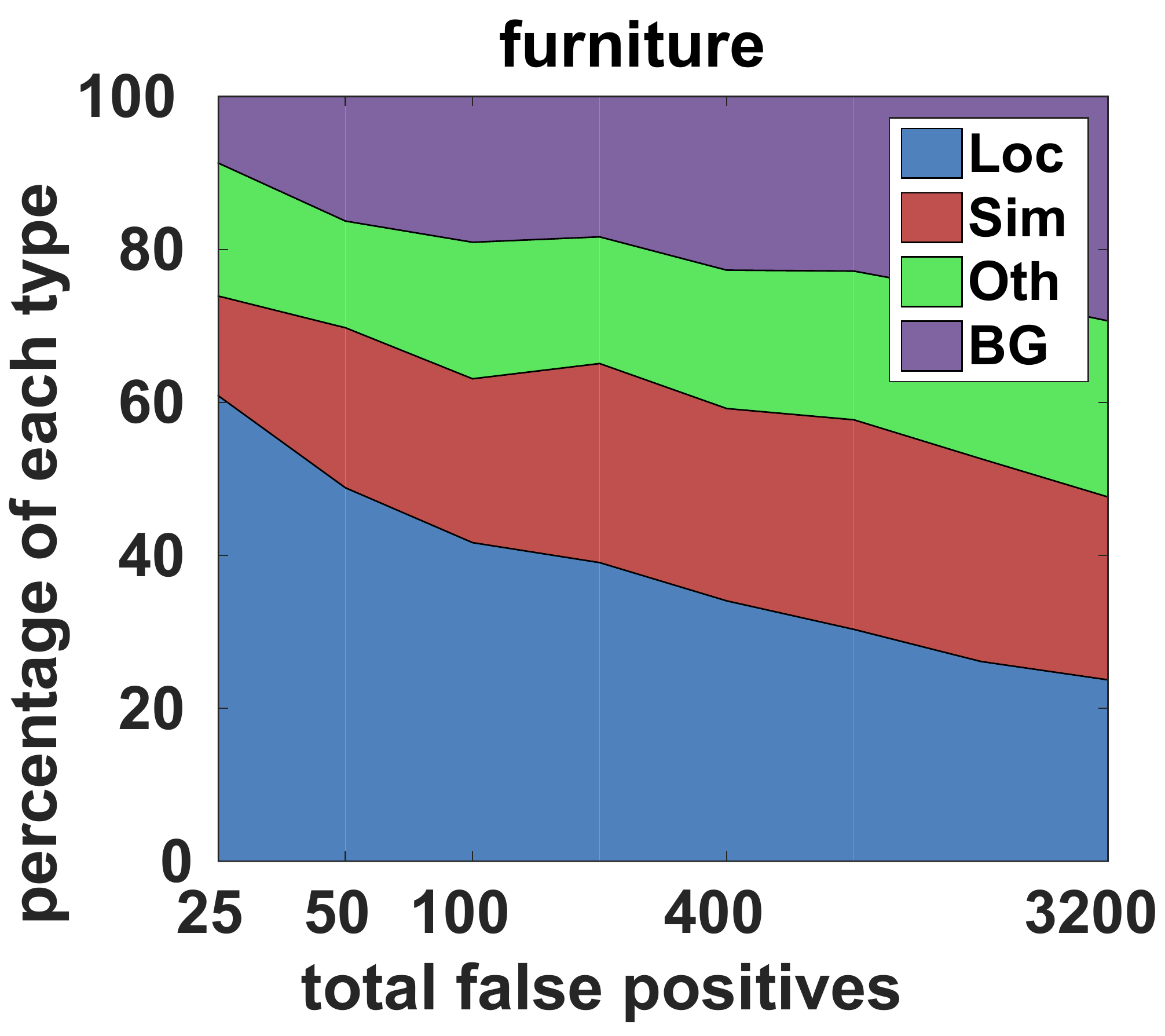}
        \caption{1Step-Grid}
    \end{subfigure}
    \caption{\small The distribution of top-ranked types of false positives (FPs). FPs are categorized into four different subcategories. The diagram shows the change in the distribution of these types when more FPs with decreasing scores are considered. \emph{Loc} represents those FPs caused by poor localization (a duplicate detection or detection with IoU between 0.1 and 0.5). \emph{Sim} shows those coming from confusion with one of the similar classes. \emph{BG} stands for FPs on background and \emph{Oth} represents other sources.} \label{fig:FP_rates}
    \vspace{-0.4cm}
 \end{figure}

To this end, we use the powerful tool of Hoeim \etal \cite{hoiem2012diagnosing}. Figure \ref{fig:FP_rates} shows the distribution of top-ranked false positive rates for \emph{G-CNN}, \emph{Fast R-CNN} and the 1Step-Grid approach defined in the previous subsection. Comparing the distributions for G-CNN and Fast R-CNN, it is clear that removing the proposal stage from the system using our method did not hurt the localization and for the furniture class, it slightly improved the FPs due to localization error. Note that 1Step-Grid is trained on the same set of training tuples as G-CNN. However, the higher rate of false positives due to localization in 1Step-Grid is another indication of the importance of G-CNN's multi-step training strategy.

\subsection{Qualitative results}
\begin{figure*}

    \centering
    \begin{subfigure}[b]{0.22\textwidth}
        \includegraphics[width=\textwidth]{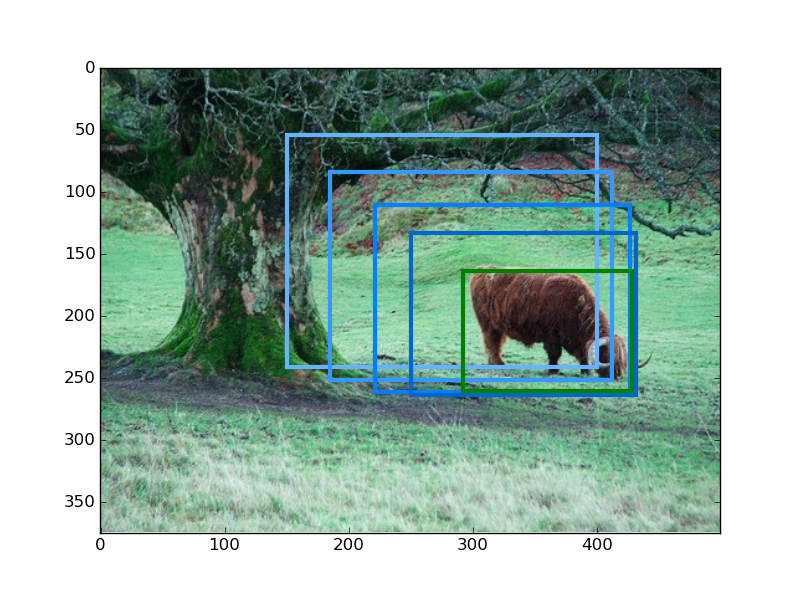}
    \end{subfigure}
    \begin{subfigure}[b]{0.22\textwidth}
        \includegraphics[width=\textwidth]{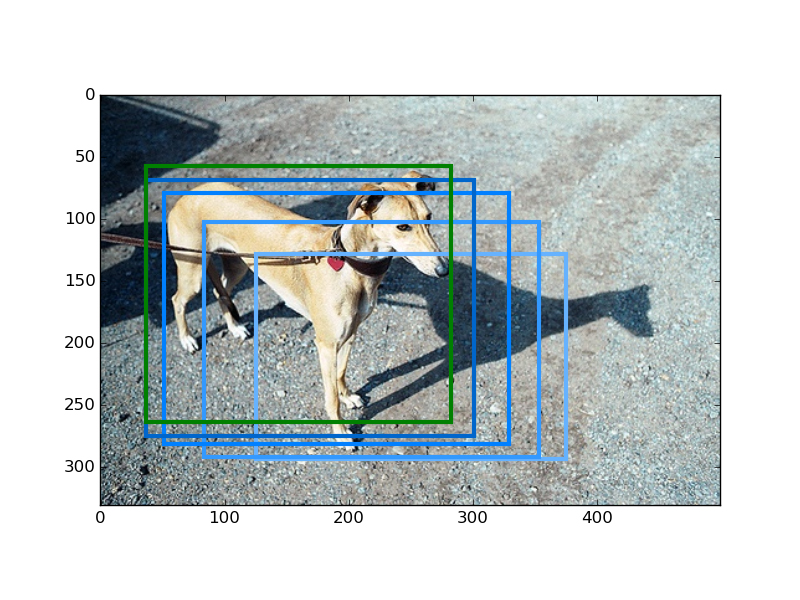}
    \end{subfigure}
    \begin{subfigure}[b]{0.22\textwidth}
        \includegraphics[width=\textwidth]{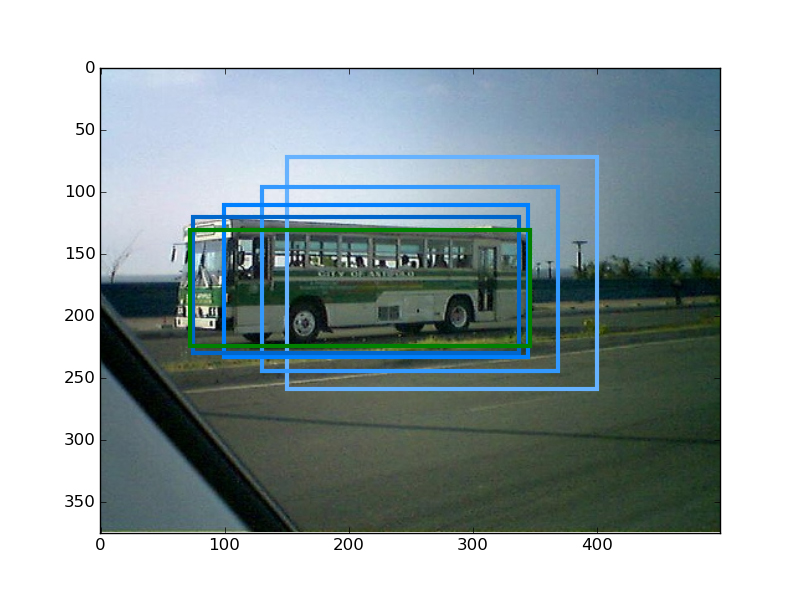}
    \end{subfigure}
    \begin{subfigure}[b]{0.22\textwidth}
        \includegraphics[width=\textwidth]{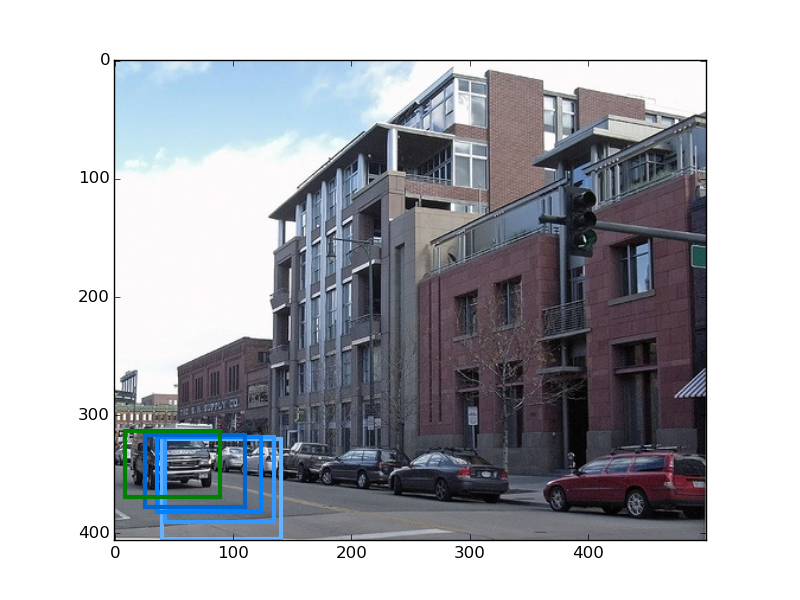}
    \end{subfigure}
    
    \centering
    \begin{subfigure}[b]{0.22\textwidth}
        \includegraphics[width=\textwidth]{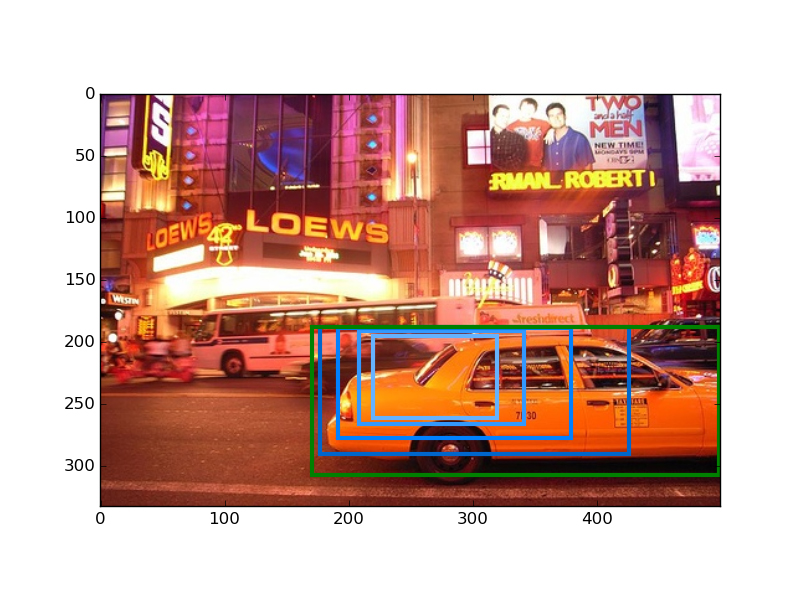}
    \end{subfigure}
    \begin{subfigure}[b]{0.22\textwidth}
        \includegraphics[width=\textwidth]{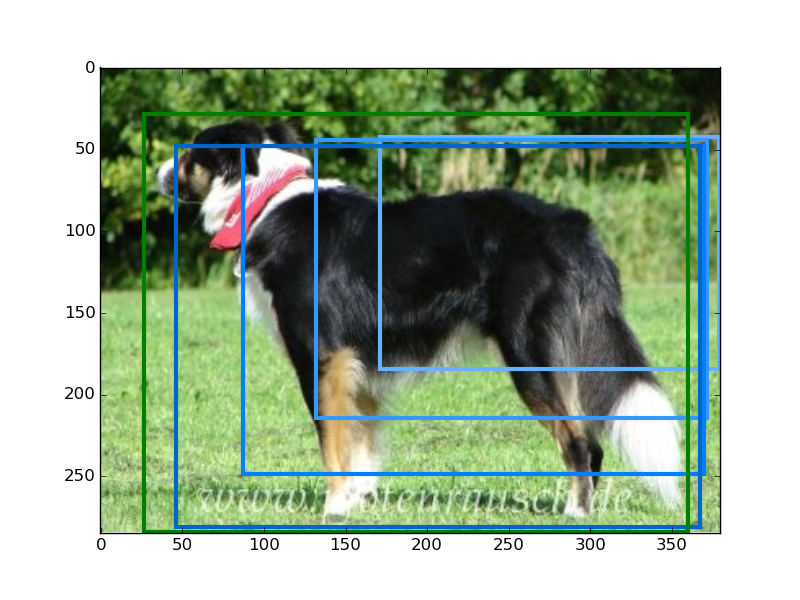}
    \end{subfigure}
    \begin{subfigure}[b]{0.22\textwidth}
        \includegraphics[width=\textwidth]{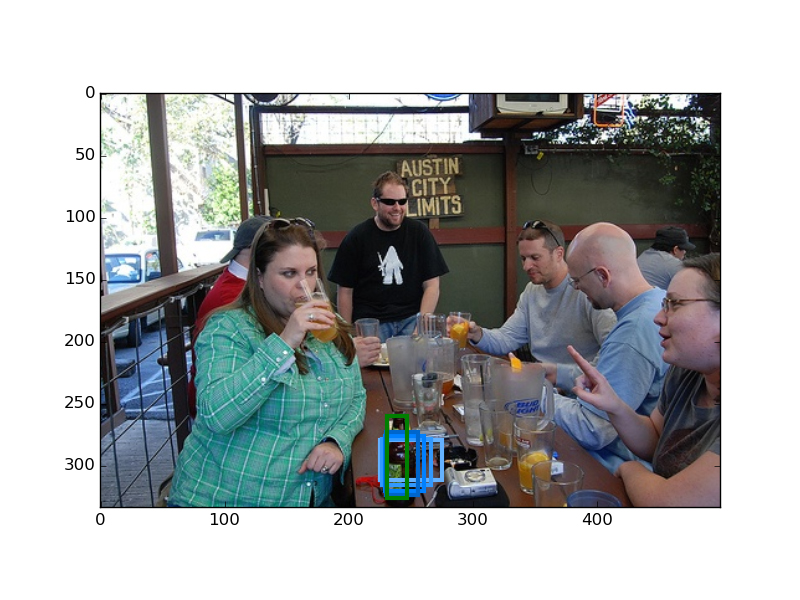}
    \end{subfigure}
    \begin{subfigure}[b]{0.22\textwidth}
        \includegraphics[width=\textwidth]{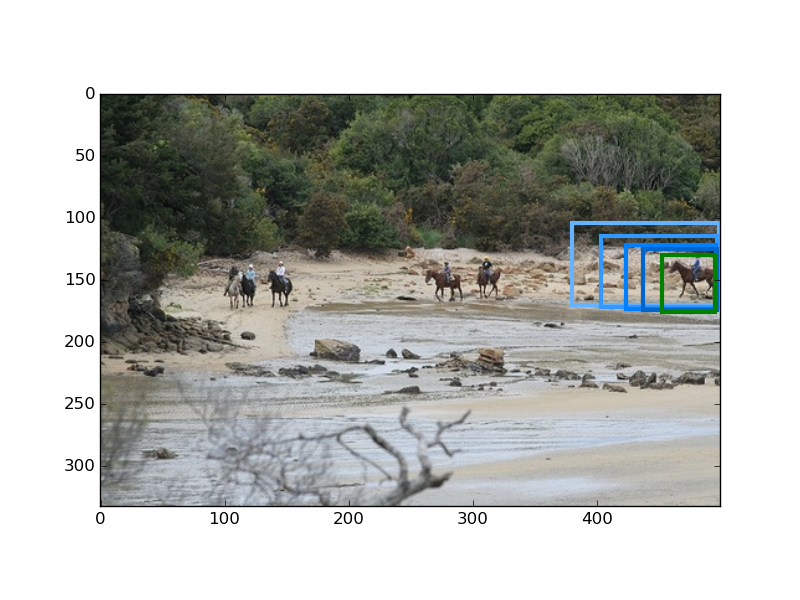}
    \end{subfigure}
    
    \centering
    \begin{subfigure}[b]{0.22\textwidth}
        \includegraphics[width=\textwidth]{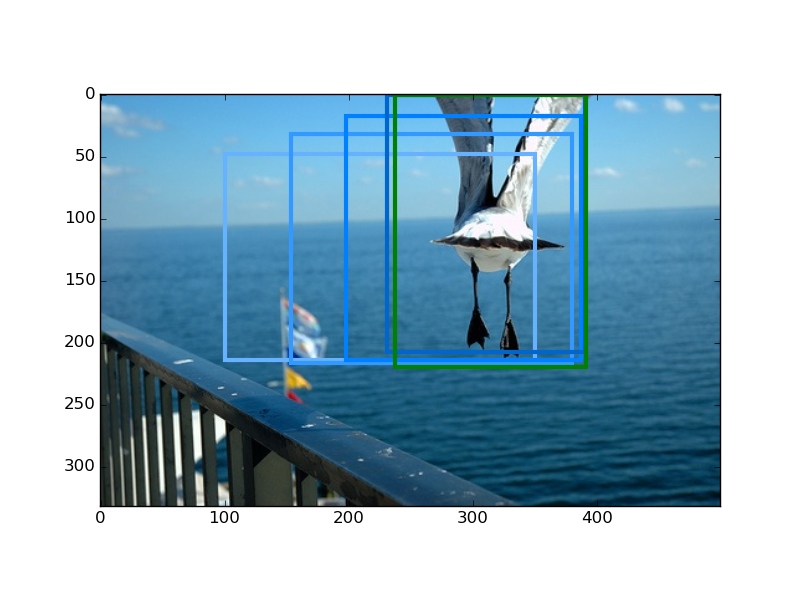}
    \end{subfigure}
    \begin{subfigure}[b]{0.22\textwidth}
        \includegraphics[width=\textwidth]{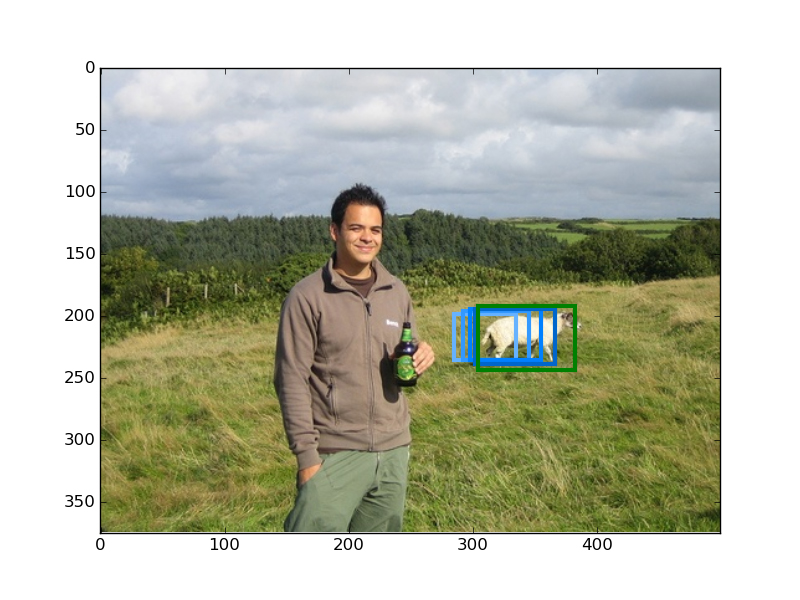}
    \end{subfigure}
    \begin{subfigure}[b]{0.22\textwidth}
        \includegraphics[width=\textwidth]{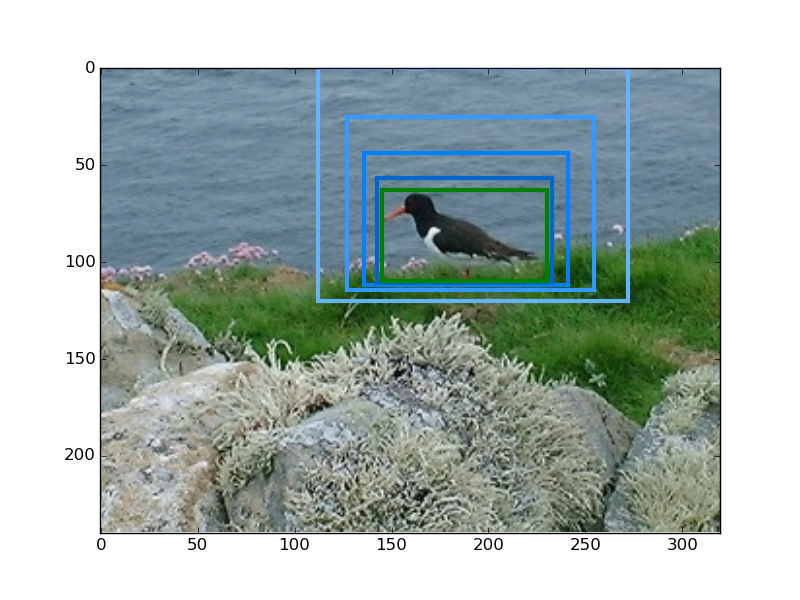}
    \end{subfigure}
    \begin{subfigure}[b]{0.22\textwidth}
        \includegraphics[width=\textwidth]{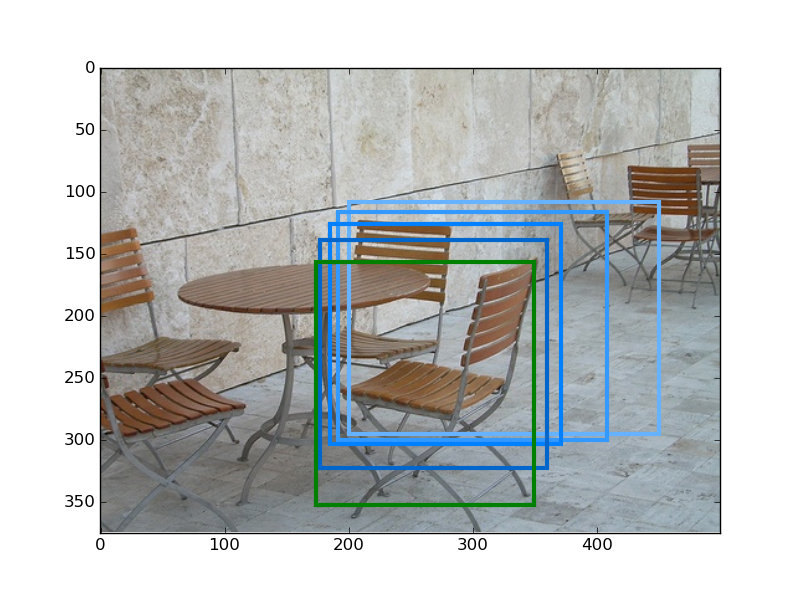}
    \end{subfigure}
    
     \begin{subfigure}[b]{0.22\textwidth}
        \includegraphics[width=\textwidth]{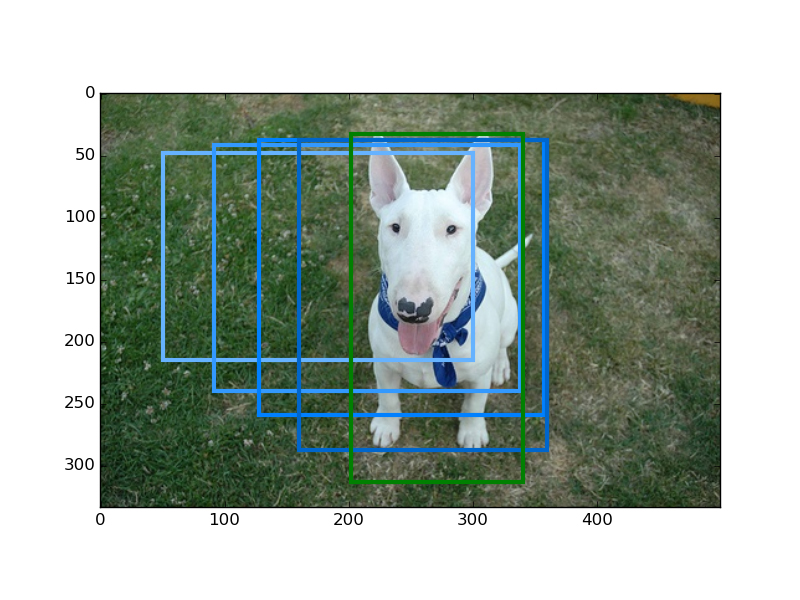}
    \end{subfigure}
     \begin{subfigure}[b]{0.22\textwidth}
        \includegraphics[width=\textwidth]{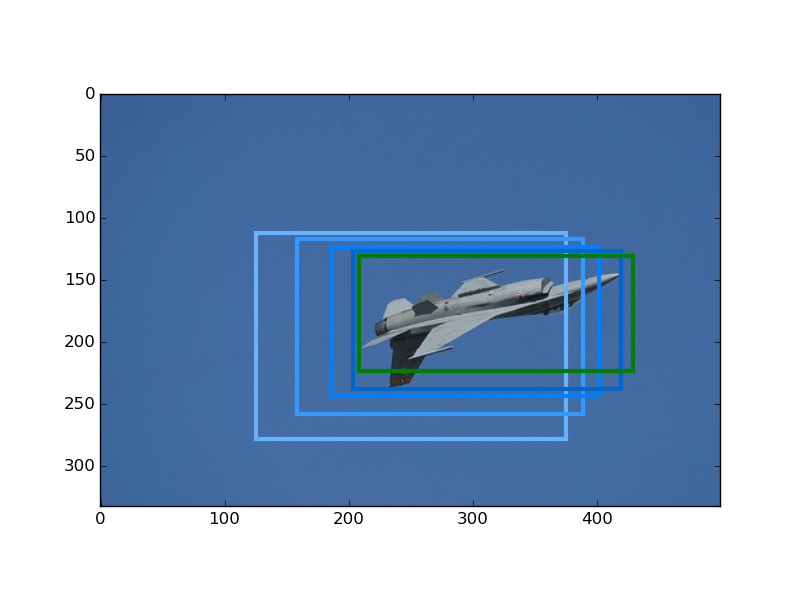}
    \end{subfigure}
     \begin{subfigure}[b]{0.22\textwidth}
        \includegraphics[width=\textwidth]{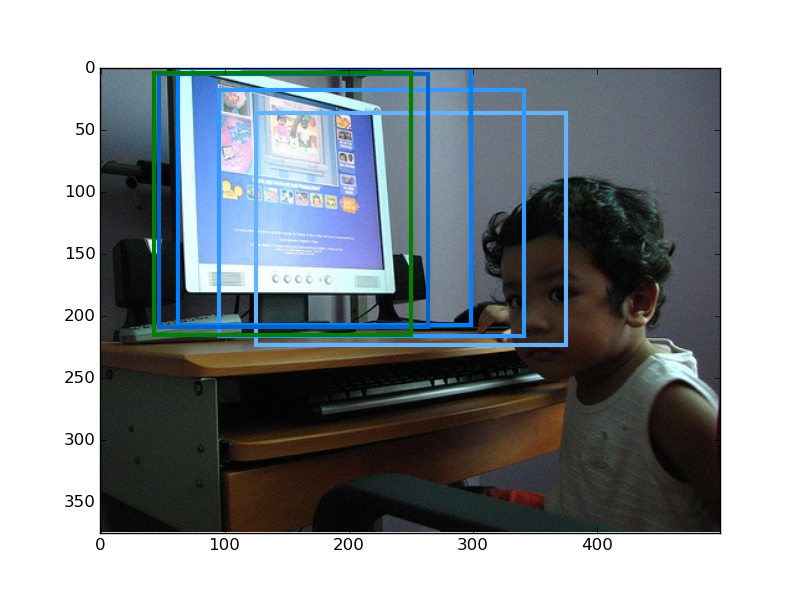}
    \end{subfigure}
     \begin{subfigure}[b]{0.22\textwidth}
        \includegraphics[width=\textwidth]{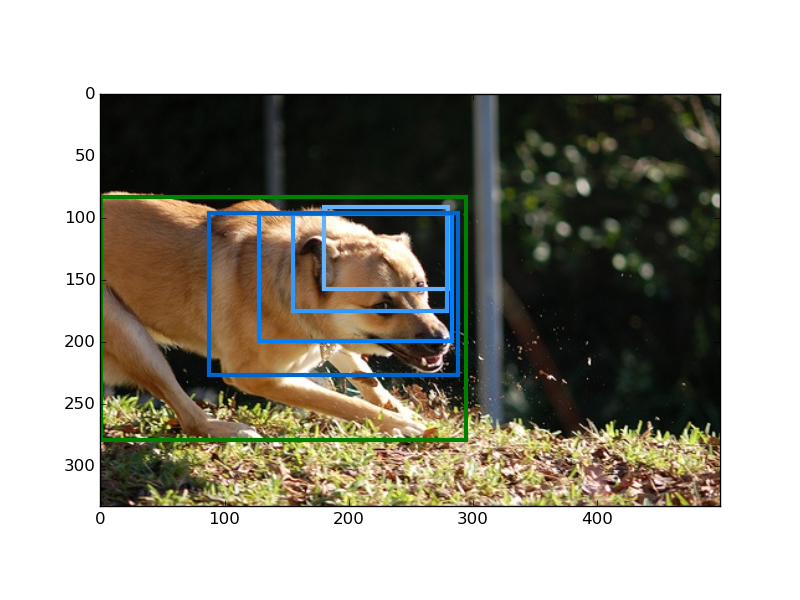}
    \end{subfigure}
    
    \centering
    \begin{subfigure}[b]{0.22\textwidth}
        \includegraphics[width=\textwidth]{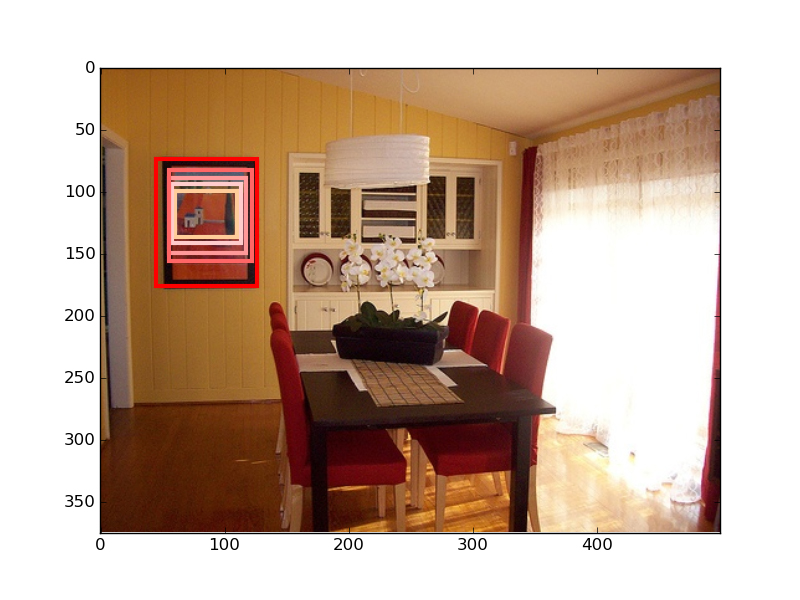}
    \end{subfigure}
    \begin{subfigure}[b]{0.22\textwidth}
        \includegraphics[width=\textwidth]{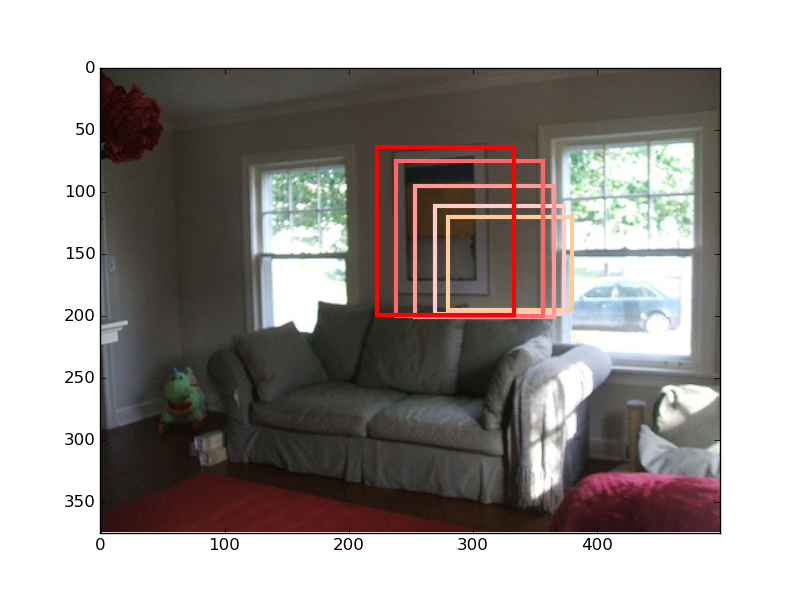}
    \end{subfigure}
    \begin{subfigure}[b]{0.22\textwidth}
        \includegraphics[width=\textwidth]{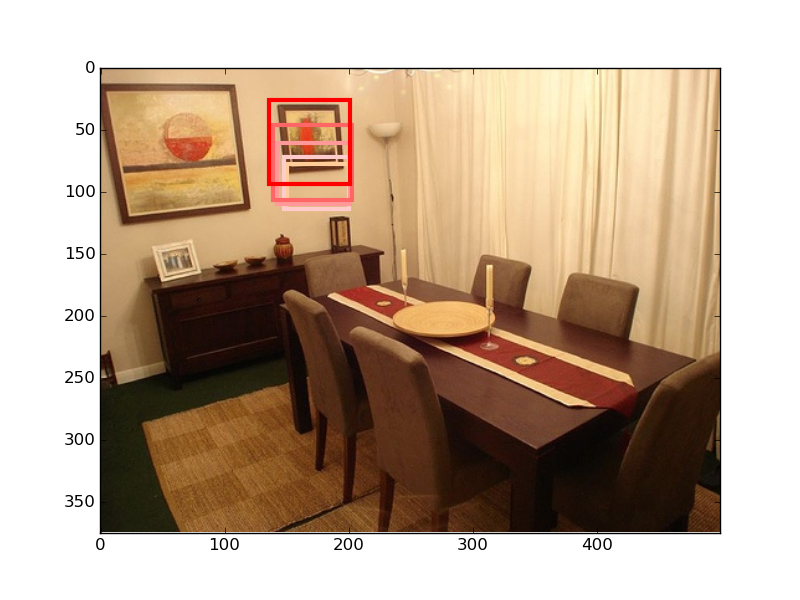}
    \end{subfigure}
    \begin{subfigure}[b]{0.22\textwidth}
        \includegraphics[width=\textwidth]{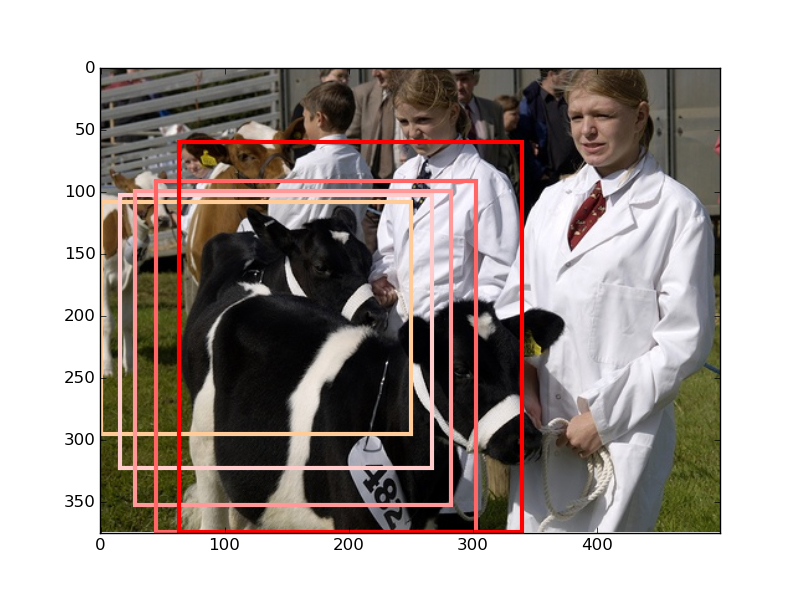}
    \end{subfigure}
      \begin{subfigure}[b]{0.22\textwidth}
        \includegraphics[width=\textwidth]{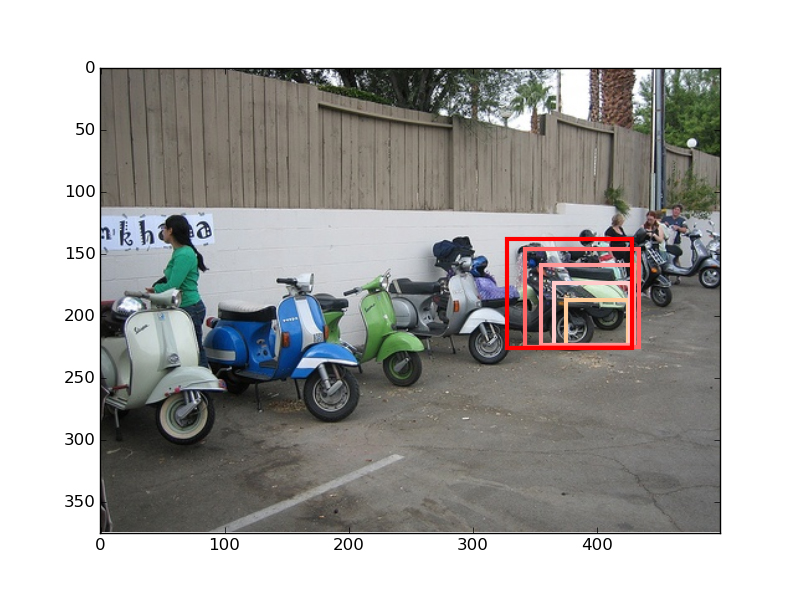}
    \end{subfigure}
      \begin{subfigure}[b]{0.22\textwidth}
        \includegraphics[width=\textwidth]{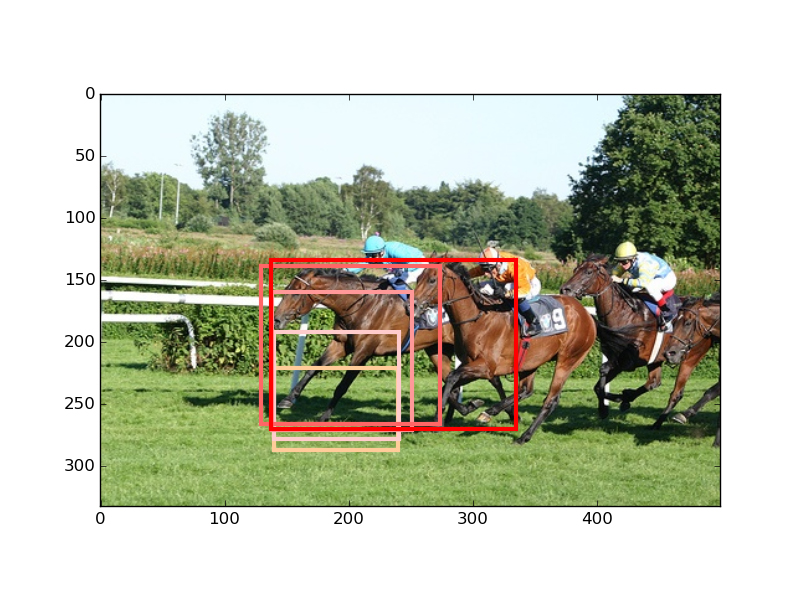}
    \end{subfigure}
      \begin{subfigure}[b]{0.22\textwidth}
        \includegraphics[width=\textwidth]{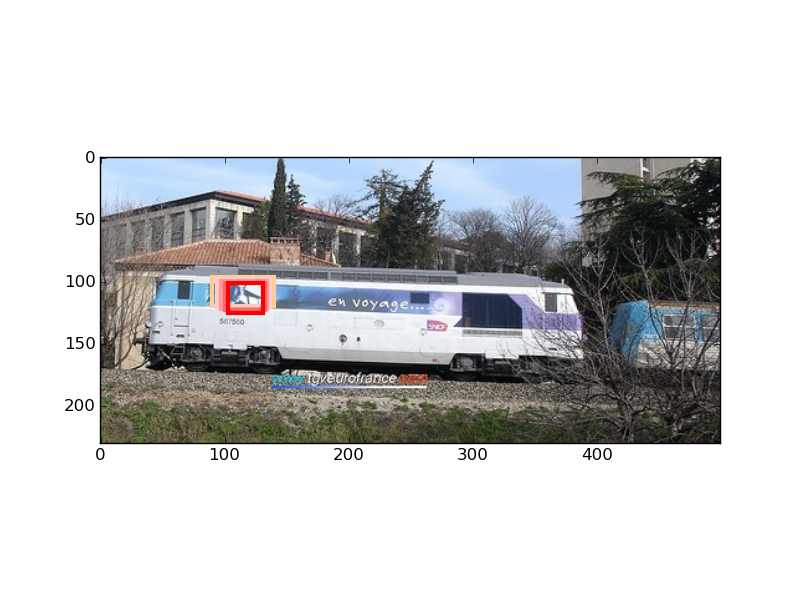}
    \end{subfigure}
      \begin{subfigure}[b]{0.22\textwidth}
        \includegraphics[width=\textwidth]{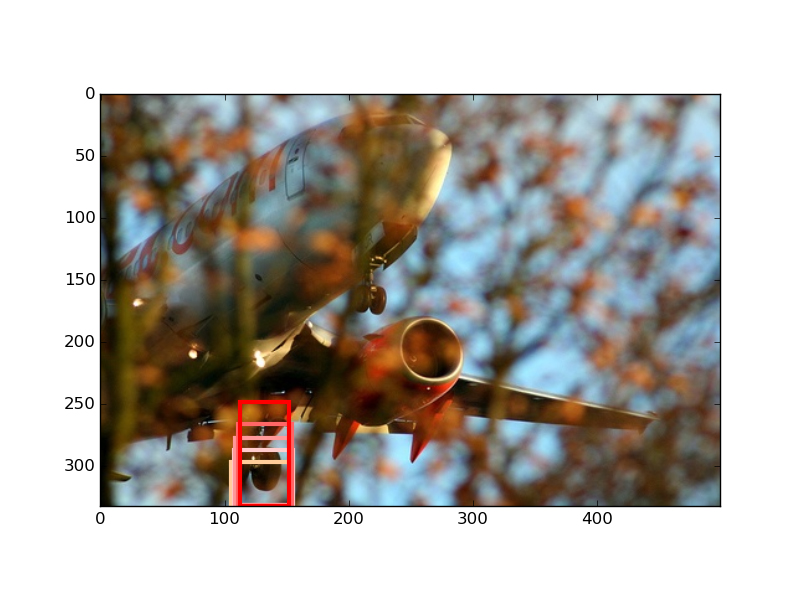}
    \end{subfigure}
    
      \begin{subfigure}[b]{0.22\textwidth}
        \includegraphics[width=\textwidth]{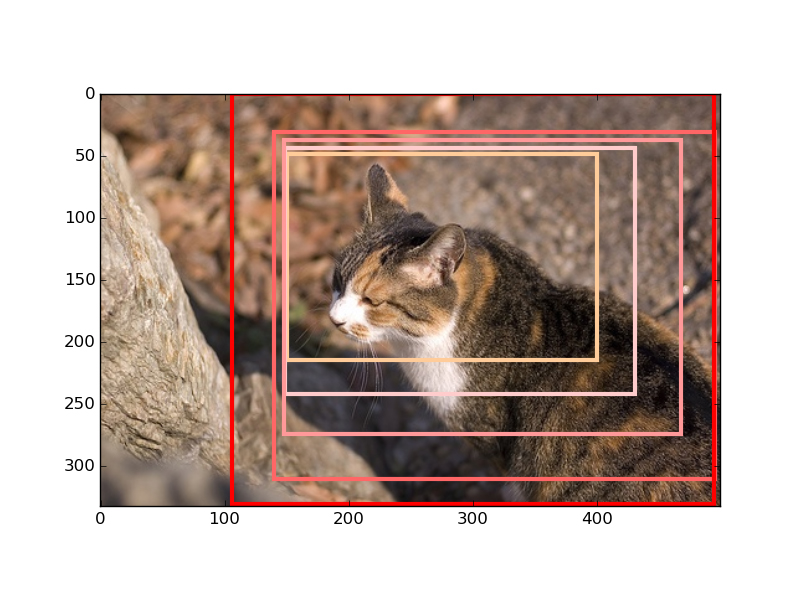}
    \end{subfigure}
      \begin{subfigure}[b]{0.22\textwidth}
        \includegraphics[width=\textwidth]{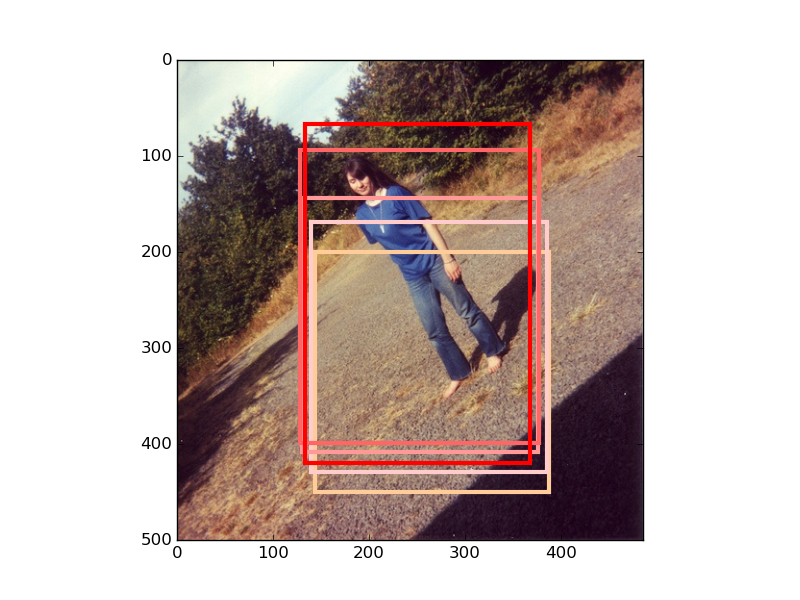}
    \end{subfigure}
      \begin{subfigure}[b]{0.22\textwidth}
        \includegraphics[width =\textwidth]{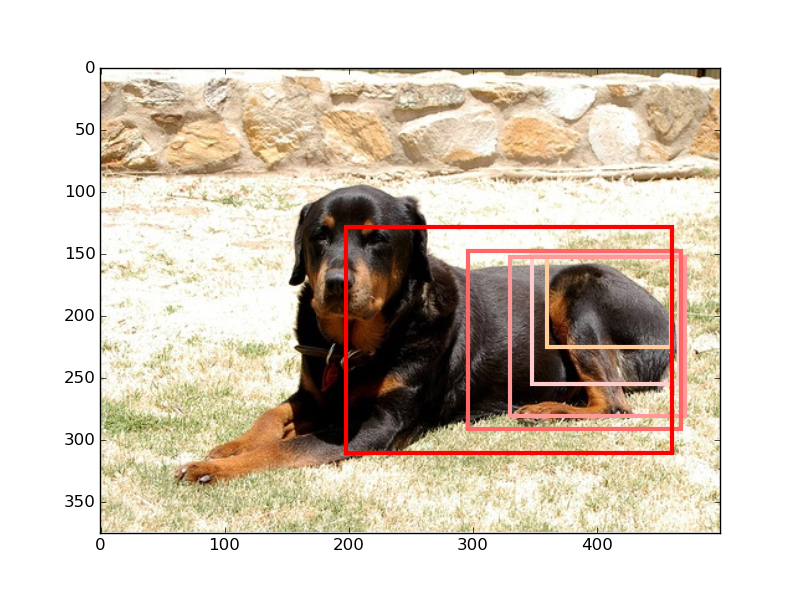}
        \end{subfigure}
      \begin{subfigure}[b]{0.22\textwidth}
        \includegraphics[width=\textwidth]{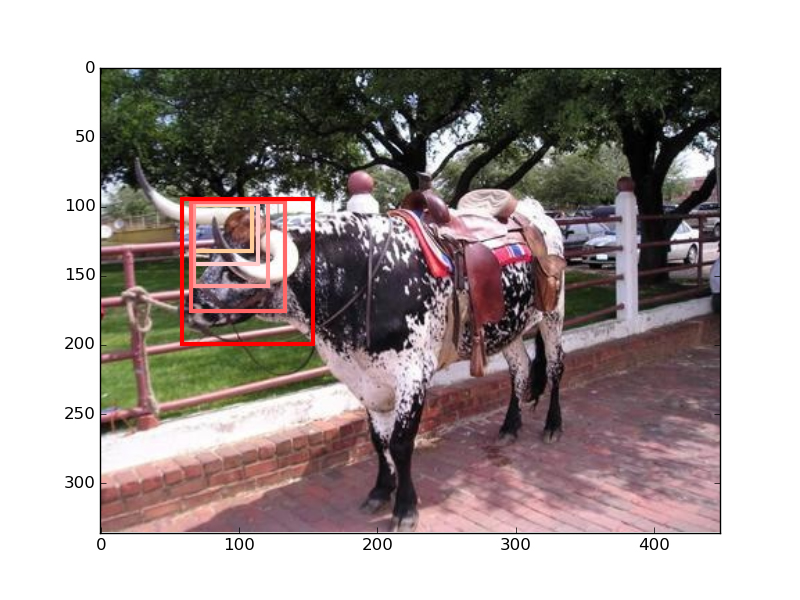}
    \end{subfigure}
    \caption{\small A sample of paths G-CNN found towards objects in the VOC2007 test set using AlexNet CNN structure. The first four rows show some success examples while the last rows show some failure cases. The most common failures of G-CNN can be categorized into the following sub-categories: false firing of the classifier on similar objects (first three failure cases in the fifth row where G-CNN fits into picture frames instead of monitors); bad localization due to similar objects with high overlaps (next three examples); false firing of the classifier on small boxes (last two cases in the sixth row); localization error due to hard pose of the object or small initial box compared to the actual size of the object (examples in the last row)}
    \label{fig:Qulitative_Results}
 \end{figure*}

Figure \ref{fig:Qulitative_Results} shows some of the paths found by G-CNN in the space of bounding boxes starting from an initial grid with three scales. This example shows how G-CNN is capable of changing the position and scale of the boxes to fit them to different objects. The first four rows show successful examples while the last ones show failure examples. 

\subsection{Detection run time}
Here we compare the detection time of our algorithm with Fast R-CNN. For both methods, we used the truncated SVD technique proposed in \cite{girshick2015fast} and compressed fc6 and fc7 layers by keeping their top 1024 singular values and 256 singular values respectively. Timings are performed on a system with two K40 GPUs. The VGG16 network structure is used for both detection techniques and G-CNN uses the same classifier as Fast R-CNN.

We used Selective Search proposal to generate around 2K bounding boxes as suggested by \cite{girshick2015fast}. This stage takes $1830$ ms to complete on average (selective search algorithm is not implemented in GPU mode). Fast R-CNN itself takes $220$ ms on average for detecting objects. This leads to a total detection time of $2050$ ms/im.

On the other hand, G-CNN does not need any object proposal stage. However, it iterates S=5 times with a grid of around 180 boxes. The global part of the network (See \ref{sec:gcnn_test}) takes $188$ ms for each image. Each iteration of the segmentation network takes $35$ ms. The classification network can be run in parallel. This would lead to a detection time of $363$ ms/im (around $3$ fps) in total.

\section{Conclusion}
We proposed G-CNN, a CNN-based object detection technique which models the problem of object detection as an iterative search in the space of all possible bounding boxes. Our model starts from a grid of fixed boxes regardless of the image content and migrates them to objects in the image. Since this search problem is nonlinear, we proposed a piece-wise regression model that iteratively moves boxes towards objects step by step. We showed how to learn the CNN architecture in a stepwise manner. The main contribution of the proposed technique is removing the object proposal stage from the detection system, which is the current bottleneck for CNN-based detection systems. G-CNN is $5X$ faster than "Fast R-CNN" and achieves comparable results to state-of-the-art detectors.

\vspace{0.25cm}

\acknowledge{Acknowledgment:}
This work was partially supported by grant N00014-10-1-0934 from ONR.

\clearpage
{\small
\bibliographystyle{ieee}
\bibliography{main}
}

\end{document}